\documentclass{article}

\usepackage{microtype}
\usepackage{graphicx}
\usepackage{subcaption}
\usepackage{booktabs} 

\usepackage{hyperref}

\usepackage[preprint]{icml2026}

\makeatletter
\icml@noticeprintedtrue
\makeatother

\usepackage{amsmath}
\usepackage{amssymb}
\usepackage{mathtools}
\usepackage{amsthm}
\usepackage{bm}

\usepackage[capitalize,noabbrev]{cleveref}

\theoremstyle{plain}

\theoremstyle{definition}

\theoremstyle{remark}

\usepackage[textsize=tiny]{todonotes}

\usepackage{enumitem}
\usepackage{booktabs}
\usepackage{multirow}
\usepackage{graphicx}

\newcommand{\papertitle}{Laplacian Frequency Hierarchies for Efficient 3D Gaussian Splatting Training}

\usepackage[table]{xcolor}
\definecolor{tabfirst}{rgb}{1, 0.6, 0.6}   
\definecolor{tabsecond}{rgb}{1, 0.8, 0.5}  
\definecolor{tabthird}{rgb}{1, 1, 0.6}     

\newcommand{\best}[1]{\cellcolor{tabfirst}\textbf{#1}}
\newcommand{\second}[1]{\cellcolor{tabsecond}#1}
\newcommand{\third}[1]{\cellcolor{tabthird}#1}

\definecolor{lightred}{rgb}{1, 0.6, 0.6}
\definecolor{lightorange}{rgb}{1, 0.8, 0.5}
\definecolor{lightyellow}{rgb}{1, 1, 0.6}

\icmltitlerunning{\papertitle}

\begin{document}











\twocolumn[
  \icmltitle{\papertitle}

  \begin{center}
    \large
    Yixiong Yang\textsuperscript{1,*},
    Sisheng Zhang\textsuperscript{1,*},
    Qingsong Yan\textsuperscript{2},
    Shaohuai Shi\textsuperscript{1},
    Qiang Wang\textsuperscript{1,\textdagger}

    \vskip 0.12cm

    \normalsize
    \textsuperscript{1}Harbin Institute of Technology, Shenzhen, China,
    \textsuperscript{2}XGRIDS, China,
    \textsuperscript{*}Equal contribution,
    \textsuperscript{\textdagger}Corresponding author
  \end{center}

  \vskip 0.20in
]




\begin{abstract}
A key bottleneck in 3D Gaussian Splatting training is the continual growth of Gaussian primitives, which increases optimization cost and slows convergence, especially at high resolutions. 
We propose \textbf{Laplacian Frequency Hierarchies}, a simple yet efficient 3DGS scheme that combines Laplacian image decomposition with coarse-to-fine, frequency-staged training. 
After fitting lower-frequency structure, we archive the corresponding Gaussian field so that subsequent fields can optimize higher-frequency residuals without carrying the full primitive burden, and we compose the rendered components in the image domain via a Laplacian-style reconstruction at inference time. 
This design reduces the number of active Gaussians during training, thereby lowering optimization overhead and accelerating training. 
The proposed scheme is \emph{plug-and-play} and \emph{orthogonal} to prior 3DGS accelerations: it can be directly combined with strong backbones such as Taming-3DGS and FastGS to improve training speed with competitive reconstruction quality. 
It achieves average training speedups of $1.73\times$ and $1.21\times$ at \emph{1K}, and $1.74\times$ and $1.33\times$ at \emph{4K} on Taming-3DGS and FastGS, with larger gains on more challenging scenes and increasingly pronounced benefits at higher resolutions. The project page and source code are available at \url{https://sorenzhang574.github.io/Laplacian-GS/}.

\end{abstract}

\section{Introduction}

\begin{figure*}[t]
    \centering
    \includegraphics[width=1.0\linewidth]{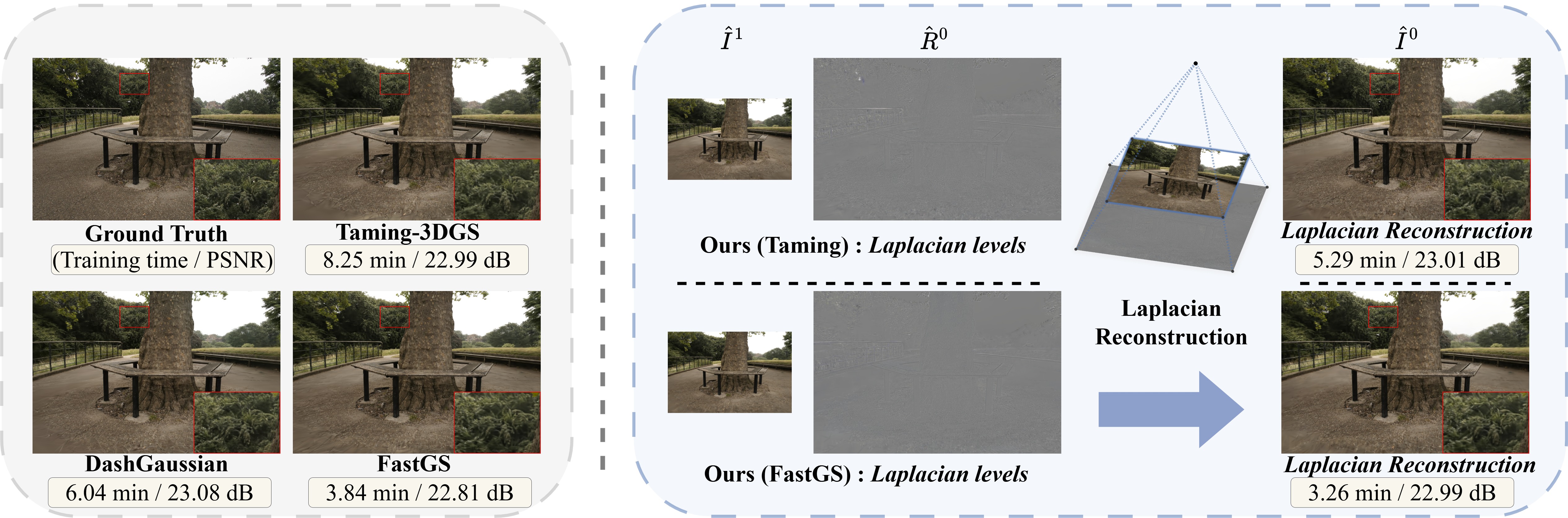}
    \vspace{-10pt}
    \caption{We propose Laplacian Frequency Hierarchies, a frequency-factorized scheme for efficient 3D Gaussian Splatting training. We adopt a Laplacian image decomposition to split supervision into a low-frequency base and high-frequency residuals, and train a sequence of smaller, frequency-specific Gaussian fields. The framework is plug-and-play with strong accelerated backbones such as Taming-3DGS \cite{mallick2024taming} and FastGS \cite{ren2025fastgs}, yielding faster optimization while maintaining comparable reconstruction quality.}
   \label{fig:teaser}
    \vspace{-12pt}
\end{figure*}

Novel view synthesis (NVS) is a foundational problem in 3D reconstruction and scene representation, aiming to render photorealistic views of a scene from previously unseen camera poses. Starting from Neural Radiance Fields (NeRF) \citep{mildenhall2021nerf}, implicit representations have achieved impressive fidelity but often at the cost of expensive optimization and slow rendering. In contrast, 3D Gaussian Splatting (3DGS) \citep{kerbl3Dgaussians} introduces an explicit primitive-based representation with differentiable rasterization, dramatically improving training and rendering efficiency, reducing optimization time from days to minutes while enabling real-time rendering. Owing to its favorable quality--speed balance, 3DGS has quickly become a backbone for a broad range of downstream tasks \citep{chen2025survey3dgaussiansplatting,Huang2DGS2024,zuo2025gaussianworld,huang2024gaussianformer,matsuki2024gaussian,guo2025articulatedgs}.

Despite these advances, efficiency in practical applications remains a bottleneck. 
As scene complexity and target resolution increase, 3DGS typically requires more Gaussian primitives and incurs heavier Gaussian--tile interactions, increasing computation and peak memory usage. 
This has motivated acceleration methods based on rasterization optimization, scheduling, and primitive control. 
Taming 3DGS \cite{mallick2024taming} improves tile-level parallelization, while DashGaussian \cite{chen2025dashgaussian} uses resolution and primitive schedulers to delay expensive phases. 
Mini-Splatting \citep{fang2024mini} and FastGS \citep{ren2025fastgs} reduce training cost by controlling densification, pruning, or adaptive density. 
Taken together, these approaches aim to reduce the average Gaussian workload during training, e.g., by shrinking the Gaussian set, reducing Gaussian–tile pairs, or postponing expensive iterations.

However, many existing approaches do not fully eliminate a structural inefficiency. They still rely on a single Gaussian set to cover both coarse and fine image components across the entire optimization. Frequency-based methods \citep{chen2025dashgaussian, nguyen2025coarse, zhang2024fregs, zeng2025frequency} typically use frequency cues to schedule supervision, for instance by changing the supervision order or adjusting the resolution. This changes the training curriculum but does not fundamentally decouple the representation, so the rendering workload can still grow with scene complexity and resolution. Crucially, coarse components that converge early are still carried through later training, leading to \textit{avoidable overhead} when the optimization focuses on high-frequency details. This observation also resonates with recent explorations of Laplacian decomposition in Gaussian-based representations. Frequency hierarchies~\citep{zhu2025large} have shown promising gains in speed and quality for 2D Gaussian image representation~\citep{zhang2024gaussianimage}, while Lavi et al.~\citep{lavi2025frequencyawaregaussiansplattingdecomposition} explored Laplacian decomposition in 3DGS for interpretability and artistic applications. Yet the role of Laplacian factorization as a practical strategy for improving \emph{training efficiency} in 3DGS remains underexplored. This motivates us to revisit Laplacian factorization as a mechanism for reducing the active optimization burden during training.

\begin{figure}[t]
    \centering
    \includegraphics[width=1.0\linewidth]{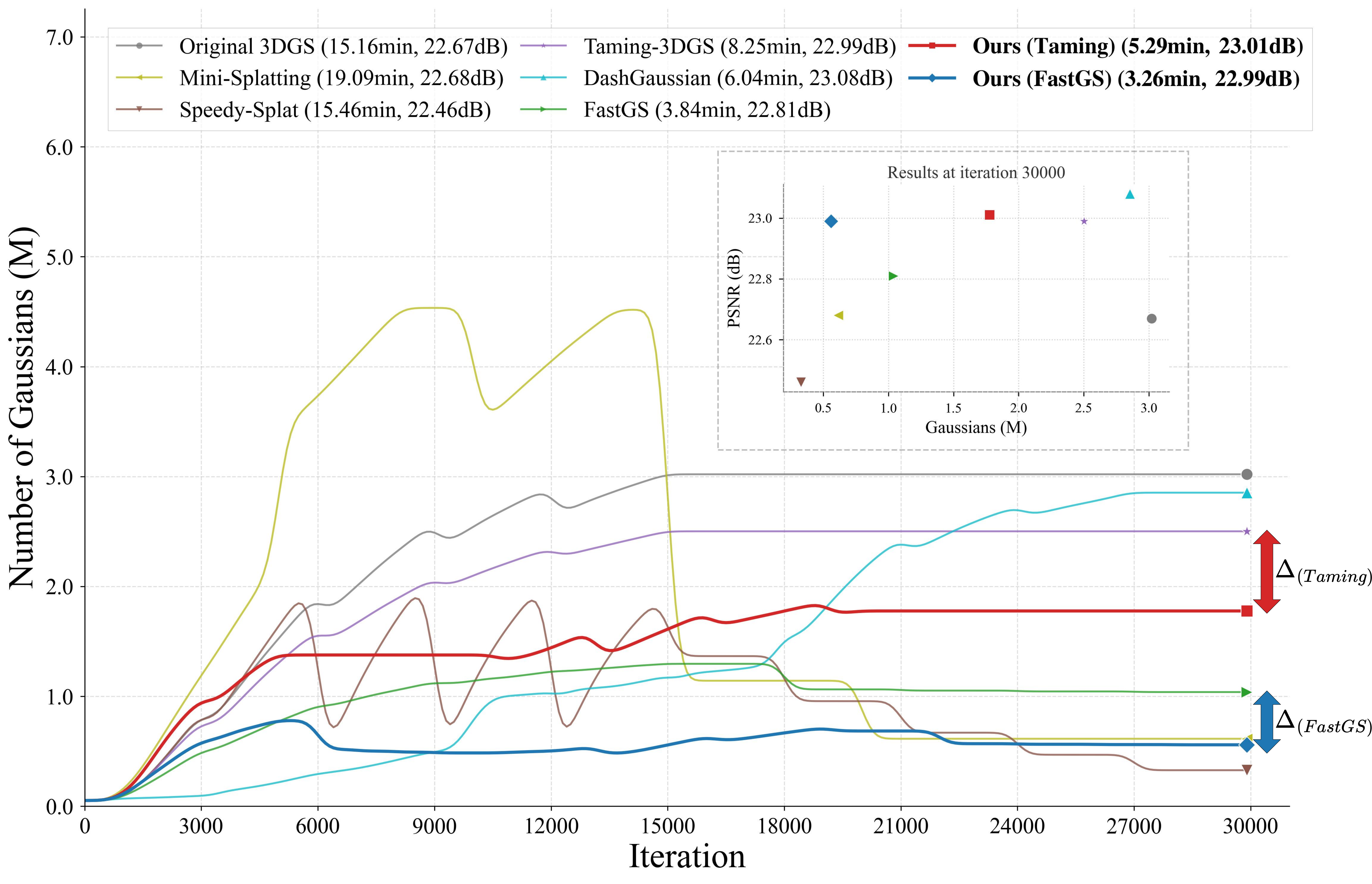}
    \vspace{-6mm}
    \caption{Number of Gaussians across training iterations on the 'treehill' scene of Mip-NeRF 360 \citep{barron2022mip}. Since earlier stages are archived, our method maintains substantially fewer \emph{active} Gaussians in later iterations.}
    \vspace{-6mm}

    \label{fig:number of gaussians}
\end{figure}

In this work, we propose \textbf{Laplacian Frequency Hierarchies}, a frequency-decomposed framework for efficient 3D Gaussian Splatting. Our key observation is that NVS ultimately produces an image-plane signal with naturally structured frequency components. Low-frequency components capture global appearance and illumination, while high-frequency components primarily encode edges, textures, and fine details. Instead of forcing one Gaussian set to bear all frequencies at all times, we perform a Laplacian decomposition of the supervision signal and reformulate 3DGS as a \emph{hierarchy} consisting of (i) a \emph{low-frequency base} Gaussian field and (ii) multiple \emph{high-frequency residual} Gaussian fields. Novel-view synthesis is then obtained by rendering each level and \emph{composing} the final image via an image-domain summation of the base rendering and residual renderings (see Fig. \ref{fig:teaser}). This decomposition naturally enables a \emph{frequency-staged training} scheme. We train the hierarchy from coarse to fine and archive each lower-frequency field once its training stage finishes. Each subsequent stage then focuses on the next frequency band without carrying the full primitive burden. Crucially, only the Gaussians of the current band are actively optimized at a time, leading to fewer active Gaussians during later iterations (Fig.~\ref{fig:number of gaussians}) and reducing overall training workload. Beyond efficiency, this Laplacian-style factorization provides an interpretable view of the representation across frequencies and can support dynamic level-of-detail (LOD) rendering and frequency-aware artistic filtering, as explored in prior work \citep{lavi2025frequencyawaregaussiansplattingdecomposition}.

We demonstrate that Laplacian Frequency Hierarchies are \emph{plug-and-play} and \emph{orthogonal} to existing 3DGS accelerations across multiple backbones. 
Our framework does not alter the core rasterizer or density-control designs, and can be directly stacked onto strong implementations such as Taming-3DGS \citep{mallick2024taming} and FastGS \cite{ren2025fastgs}. 
At \emph{1K} resolution, our framework matches state-of-the-art reconstruction quality while providing a noticeable efficiency gain. Notably, the efficiency gains become more pronounced at higher resolutions. On \emph{2K} and \emph{4K} scenes, our method delivers larger speedups while maintaining comparable reconstruction quality, demonstrating strong scalability to high-resolution settings.
Overall, our results suggest that frequency decomposition can serve as an effective training and optimization strategy for reducing the active Gaussian workload in 3DGS training. In summary, we make the following contributions:

\begin{itemize}
    \item We propose \textbf{Laplacian Frequency Hierarchies}, a Laplacian-inspired training scheme for 3DGS that decomposes image supervision into a low-frequency base component and high-frequency residual components, which are optimized by frequency-specific Gaussian fields and composed in the image domain for NVS.
    \item We develop a coarse-to-fine \textbf{frequency-staged training scheme} that archives earlier levels and trains each subsequent level within its corresponding frequency band, reducing training workload and peak memory usage.
    \item Our method shows \textbf{plug-and-play} compatibility across multiple 3DGS backbones. It improves training efficiency in the \emph{1K} setting while maintaining competitive rendering quality, and becomes increasingly favorable at \emph{2K} and \emph{4K}, yielding larger training speedups.
\end{itemize}

\section{Related Work}

\noindent \textbf{Novel view synthesis.} 
NVS has long been a central problem in 3D vision. 
NeRF established neural implicit representations for this task, followed by efforts to improve their efficiency \citep{mildenhall2021nerf,mueller2022instant}. 
3D Gaussian Splatting (3DGS) \citep{kerbl3Dgaussians} further advances efficient NVS with explicit Gaussian primitives and differentiable splatting, enabling GPU-friendly rasterization. 
Its efficiency has enabled adoption in large-scale reconstruction \citep{liu2025citygaussian}, avatars \citep{yang2024deformable}, surface reconstruction \citep{Guedon_2024_CVPR,wolf2024gsmesh}, world modeling \citep{zuo2025gaussianworld}, SLAM \citep{matsuki2024gaussian, Hu2025VTGSSLAM}, and digital twinning \citep{guo2025articulatedgs}.

\noindent \textbf{Acceleration for 3DGS training.}
Existing efforts to accelerate 3DGS training can be broadly grouped into several directions.
\emph{(i) Backpropagation and optimizers.}
Taming 3DGS \citep{mallick2024taming} replaces per-pixel backpropagation with a per-splat parallelization scheme, yielding substantial speedups with strong generality.
Other works revisit optimizers and update rules, such as 3DGS-LM \citep{hoellein_2025_3dgslm}, which replaces Adam with Levenberg--Marquardt to accelerate convergence, and 3DGS$^{2}$ \citep{lan20253dgs2}, which explores second-order optimization for faster training.
\emph{(ii) Scheduling and frequency-aware training.}
A related line improves efficiency by progressively changing the training signal or workload, including resolution or frequency scheduling and coarse-to-fine strategies \citep{chen2025dashgaussian, farooq2025optimized}. DashGaussian \citep{chen2025dashgaussian} introduces resolution and primitive schedulers that defer rapid primitive growth to later iterations, reducing the total training cost.
\emph{(iii) Densification control and pruning.}
Many methods explicitly manage densification and pruning to curb the growth of Gaussian primitives.
Mini-Splatting \citep{fang2024mini} removes a large portion of Gaussians via simplification based on intersection preservation and sampling.
Speedy-Splat \citep{HansonSpeedy} prunes Gaussians using a Hessian approximation aggregated across training views.
FastGS \citep{ren2025fastgs} designs densification and pruning rules based on multi-view consistency and achieves strong efficiency--quality trade-offs.
\emph{(iv) System and parallelization.}
System-oriented efforts explore parallel and distributed training \citep{zhao2024scaling3dgaussiansplatting} as well as deployment-oriented designs for resource-constrained devices.
Orthogonally, high-resolution settings introduce additional scalability challenges, motivating methods tailored for large images, such as \citep{liu2025efficientgs, dhiman2024turbo, li2025hrgs}. 
Our method is most closely related to frequency-based approaches. Unlike prior work that primarily uses frequency for scheduling or reweighting, we incorporate frequency at the representation level by \emph{decomposing} the 3DGS scene into frequency-specific fields and \emph{composing} their renderings in the image domain via a Laplacian-style reconstruction. This improves training efficiency and scales better to high-resolution settings.

\noindent \textbf{Frequency-based methods in 3DGS. }Several recent works exploit frequency or multi-scale priors to improve 3DGS training and rendering. FreGS \citep{zhang2024fregs} regularizes 3DGS in the Fourier domain to improve NVS fidelity and mitigate over-reconstruction artifacts. Opti3DGS and AutoOpti3DGS \citep{farooq2025optimized,nguyen2025coarse} employ frequency-aware coarse-to-fine supervision while retaining a single Gaussian field. Zeng et al. \citep{zeng2025frequency} introduce frequency-aware densification and deletion to allocate finer Gaussians to high-frequency regions, improving fidelity with fewer Gaussians.
Lavi et al.~\citep{lavi2025frequencyawaregaussiansplattingdecomposition} use Laplacian subbands for LOD rendering and frequency-aware editing.
In contrast, we factorize the scene into frequency-specific Gaussian fields and archive completed fields, so later stages optimize only the active frequency band, directly reducing the optimization workload for faster training.

\section{Methodology}

\begin{figure*}[!t]
    \centering
    \includegraphics[width=0.85\linewidth]{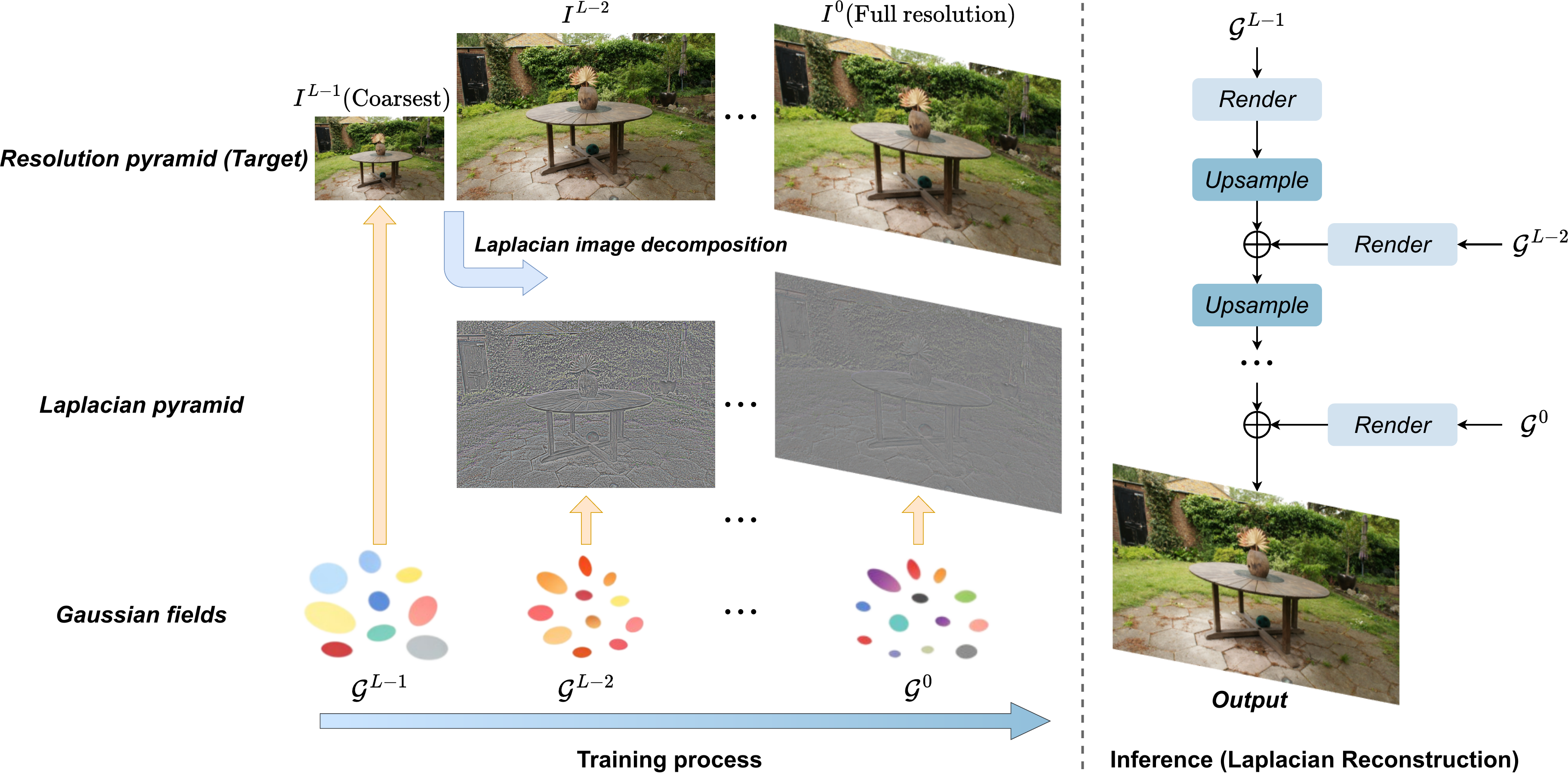}
    \vspace{-1mm}
    \caption{\textbf{Overview of our method.} We decompose each training image into a resolution pyramid and the corresponding Laplacian residual pyramid. During training, a sequence of Gaussian fields $\mathcal{G}^{L-1}, \mathcal{G}^{L-2}, \ldots, \mathcal{G}^{0}$ is optimized in a coarse-to-fine manner, with each field learning a specific frequency component. At inference time, the fields are rendered separately and their outputs are composed via Laplacian reconstruction to produce the full-resolution result.}
    \vspace{-4mm}

    \label{fig:method}
\end{figure*}

\subsection{Preliminaries.}
3D Gaussian Splatting (3DGS) represents a scene using a set of anisotropic Gaussian primitives
$\mathcal{G}=\{G_i\}_{i=1}^{N}$ \citep{kerbl3Dgaussians}.
Each primitive is parameterized by a 3D mean $\mu_i\in\mathbb{R}^3$, an opacity $\sigma_i\in(0,1)$,
a covariance matrix $\Sigma_i\in\mathbb{R}^{3\times 3}$ (e.g., $\Sigma_i = R_i S_i S_i^{\top} R_i^{\top}$ with
rotation $R_i$ and diagonal scale $S_i=\mathrm{diag}(s_i)$), and view-dependent appearance parameters.
The Gaussian density is
\begin{equation}
G_i(x)=\exp\!\left(-\frac{1}{2}(x-\mu_i)^{\top}\Sigma_i^{-1}(x-\mu_i)\right)
\end{equation}

To render an image from a camera view, 3DGS projects each 3D Gaussian onto the image plane using an affine
approximation of the perspective projection, yielding an elliptical 2D footprint whose contribution at pixel coordinate $p$ is weighted by a 2D Gaussian kernel $G_i^{2D}(p)$.
Each primitive contributes an alpha value $\alpha_i(p)=\sigma_i\,G_i^{2D}(p)$ and a view-dependent color $c_i(\mathbf{d})$.
Given depth-sorted primitives along the viewing ray, the pixel color is computed by standard alpha compositing:
\begin{equation}
\hat I(p)=\sum_{i=1}^{N} T_i(p)\,\alpha_i(p)\,c_i(\mathbf{d}),
\end{equation}
where $T_i(p)=\prod_{j=1}^{i-1}\big(1-\alpha_j(p)\big)$
denotes the accumulated transmittance.
For view-dependent appearance, 3DGS models color using spherical harmonics (SH) coefficients.
Specifically, each primitive stores SH coefficients $\{a_{i,\ell m}\}$, and we evaluate
\begin{equation}
c_i(\mathbf{d})=\sum_{k=0}^{L_{\mathrm{sh}}}\sum_{m=-k}^{k} a_{i,km}\,Y_{km}(\mathbf{d}),
\end{equation}
where $\mathbf{d}$ denotes the viewing direction, $Y_{km}$ are SH basis functions, and $L_{\mathrm{sh}}$ is the maximum SH degree. 
As the number of Gaussian primitives $N$ grows and the rendering resolution increases, both the per-iteration rendering/backpropagation cost and the peak memory footprint rise substantially, which becomes a key bottleneck for efficient training.

\subsection{Laplacian Frequency Hierarchy}
Fig.~\ref{fig:method} illustrates our Laplacian frequency hierarchy.
Our goal is to reduce the training workload of 3DGS by factorizing a single scene representation into multiple frequency-specific Gaussian fields and composing their renderings in the image domain.

\noindent \textbf{Resolution pyramid and Laplacian decomposition.}
Our image-space decomposition and reconstruction follow the classical Laplacian pyramid formulation~\citep{Burt1983}.
Let $I^{0}$ denote the full-resolution ground-truth image for a training view.
We construct an $L$-level resolution pyramid $\{I^{\ell}\}_{\ell=0}^{L-1}$ with an anti-aliased downsampling operator $\mathrm{Down}(\cdot)$ (Gaussian filtering and subsampling), and define $\mathrm{Up}(\cdot)$ as the paired upsampling operator used to construct the Laplacian residual targets $\{R^{\ell}\}_{\ell=0}^{L-2}$ and to perform the subsequent reconstruction. The corresponding relations are:
\begin{equation}
\label{eq:pyr_and_res}
I^{\ell+1}=\mathrm{Down}(I^{\ell}), \qquad
R^{\ell}=I^{\ell}-\mathrm{Up}(I^{\ell+1}),
\end{equation}
where $\ell\in\{0,\ldots,L-2\}$ indexes pyramid levels (level $0$ is full resolution and larger $\ell$ is coarser).
In practice, we apply the same decomposition to every training view $v$, yielding a resolution pyramid (also referred to as a Gaussian image pyramid) $\{I^{0}_{v}, I^{1}_{v}, \ldots, I^{L-1}_{v}\}$ and Laplacian pyramids $\{R^{0}_{v}, R^{1}_{v}, \ldots, R^{L-2}_{v}, I^{L-1}_{v}\}$.

\noindent \textbf{Frequency-specific Gaussian fields.}
We represent the scene with a hierarchy of Gaussian fields
$\{\mathcal{G}^{L-1}, \ldots, \mathcal{G}^{0}\}$ aligned with the Laplacian components.
The coarsest field $\mathcal{G}^{L-1}$ models the low-frequency content at level $L-1$ (target $I^{L-1}$),
while each field $\mathcal{G}^{\ell}$ for $\ell\in\{0,\ldots,L-2\}$ models the residual band $R^{\ell}$.
A practical difference between the base and residual targets is their value range:
$I^{L-1}$ lies in $[0,1]$, whereas each residual $R^{\ell}$ is signed and lies in $[-1,1]$.
Standard 3DGS rasterizers assume non-negative color outputs and commonly apply a post-processing step to map the
predicted SH color to $[0,1]$, e.g., by shifting and clamping:
\begin{equation}
c^{\mathrm{rgb}}_i(\mathbf{d})=\mathrm{clamp}\!\big(c_i(\mathbf{d})+0.5,\,0,\,1\big).
\label{eq:rgb_clamp}
\end{equation}
To support residual learning, we modify the rasterizer for residual fields by removing the shift and clamp, so that the rendered color can take signed values and match the range.

Given a view $v$, we render each field using the corresponding splatting renderer and denote the resulting image by
\begin{equation}
\begin{aligned}
\hat I^{L-1}(v) &= \mathrm{Render}_{\mathrm{base}}(\mathcal{G}^{L-1};v), \\
\hat R^{\ell}(v) &= \mathrm{Render}_{\mathrm{res}}(\mathcal{G}^{\ell};v),
\end{aligned}
\label{eq:render_base_res}
\end{equation}
where $\mathrm{Render}_{\mathrm{base}}$ uses the standard color mapping in Eq.~\eqref{eq:rgb_clamp}, and
$\mathrm{Render}_{\mathrm{res}}$ outputs signed residuals with the modified rasterizer.

\noindent \textbf{Image-domain composition.}
We synthesize the final full-resolution image by composing the rendered coarsest level and residuals in the image
domain via a Laplacian-style reconstruction:
\begin{equation}
\label{eq:lap_recon_pred}
\hat I^{\ell}_v = \mathrm{Up}(\hat I^{\ell+1}_v) + \hat R^{\ell}_v, \qquad \ell=L-2,\ldots,0,
\end{equation}
where $\hat I^{0}_v$ is the full-resolution prediction for view $v$.

\subsection{Frequency-staged training with archiving}
Coarse-to-fine training has been widely adopted in 3DGS to improve optimization efficiency \citep{chen2025dashgaussian, farooq2025optimized}, since low-frequency structures typically converge earlier than high-frequency details.
Building upon our Laplacian decomposition, we further introduce an \emph{archiving} mechanism: once a frequency level has finished, we remove its Gaussian field from further optimization and memory residency. Unlike simple freezing, archiving allows a trained field to be moved off GPU memory (and optionally stored on CPU/disk), so that subsequent stages can dedicate computation and memory to the next frequency band. This reduces the number of Gaussians that remain active on the GPU at each stage, improving training throughput and lowering peak VRAM usage. The training procedure is summarized in Algorithm~\ref{alg:lap_hierarchy}.

\noindent \textbf{Stage-wise training objectives.}
A naive strategy is to train each field to directly match its Laplacian target, i.e.,
$\hat I^{L-1}(v)$ to $I^{L-1}_v$ and $\hat R^{\ell}(v)$ to $R^{\ell}_v$.
In practice, we find that per-level approximation errors can accumulate across levels and lead to degraded final
reconstruction quality.
To mitigate this issue, we supervise each stage using the \emph{reconstructed} image at the corresponding resolution level.
Specifically, after training the coarsest field $\mathcal{G}^{L-1}$, we proceed from coarse to fine. For each finer level $\ell\in\{L-2,\ldots,0\}$, we render the residual field to obtain $\hat R^{\ell}_v$ and reconstruct $\hat I^{\ell}_v$ using Eq.~\ref{eq:lap_recon_pred} together with the previous-level prediction $\hat I^{\ell+1}_v$. In practice, we only cache $\hat I^{\ell+1}_v$ in the dataloader, so the archived Gaussian field does not need to remain in memory. We then optimize $\mathcal{G}^{\ell}$ by comparing $\hat I^{\ell}_v$ with the ground truth at the same pyramid level $I^{\ell}_v$ using the standard 3DGS photometric loss (e.g., a weighted combination of $\ell_1$ and SSIM). This supervision anchors each stage to the correct target resolution and reduces error accumulation across levels.

{\setlength{\textfloatsep}{2pt}\setlength{\floatsep}{2pt}\setlength{\intextsep}{2pt}
\begin{algorithm}[t]
\caption{Frequency-staged Training with Archiving for Laplacian Frequency Hierarchies}
\label{alg:lap_hierarchy}
\small
\begin{algorithmic}[1]
\STATE \textbf{Precompute pyramids.}
For each training view $v$, build the resolution pyramid $\{I_v^{\ell}\}_{\ell=0}^{L-1}$.
\STATE Initialize an empty cache for accumulated predictions $\{\hat I_v^{\ell}\}$.

\FOR{$\ell=L-1,\,L-2,\,\ldots,\,0$}
    \STATE \textbf{Initialize Gaussian field.}
    \IF{$\ell=L-1$}         \STATE Initialize $\mathcal{G}^{L-1}$ from COLMAP (default 3DGS init).
    \ELSE
        \STATE Initialize $\mathcal{G}^{\ell}$ from the previous stage field $\mathcal{G}^{\ell+1}$.
    \ENDIF

    \FOR{each training iteration}
        \STATE Sample a training view $v$.
        \IF{$\ell=L-1$}
            \STATE Render $\hat I_v^{\ell} \leftarrow \mathrm{Render}_{\mathrm{base}}(\mathcal{G}^{L-1};v)$.
        \ELSE
            \STATE Load cached $\hat I_v^{\ell+1}$ and compute $\mathrm{Up}(\hat I_v^{\ell+1})$.
            \STATE Render residual $\hat R_v^{\ell} \leftarrow \mathrm{Render}_{\mathrm{res}}(\mathcal{G}^{\ell};v)$.
            \STATE Reconstruct $\hat I_v^{\ell} \leftarrow \mathrm{Up}(\hat I_v^{\ell+1}) + \hat R_v^{\ell}$.
        \ENDIF
        \STATE Compute photometric loss $\mathcal{L}_{\mathrm{photo}}\!\left(\hat I_v^{\ell},\, I_v^{\ell}\right)$.
        \STATE Backpropagate and update parameters.
        \STATE Apply densification and pruning.
    \ENDFOR
    \STATE Cache the accumulated predictions $\hat I_v^{\ell}$ for the next stage.
    \STATE Save the trained Gaussian field $\mathcal{G}^{\ell}$ and archive it.
\ENDFOR

\end{algorithmic}
\end{algorithm}
}

\noindent \textbf{Initialization and inheritance.}
For the coarsest field $\mathcal{G}^{L-1}$, we follow standard 3DGS and initialize Gaussian positions from the COLMAP point cloud. When moving from level $\ell+1$ to $\ell$, we initialize the new field $\mathcal{G}^{\ell}$ by inheriting Gaussians from the previous level. Concretely, we copy only the 3D positions while reinitializing the remaining attributes such as scale, opacity, and colors, since different frequency bands exhibit different statistics. We also introduce an inheritance ratio $\rho\in(0,1]$: when $\rho=1.0$ all Gaussians are inherited, and when $\rho<1.0$ we randomly inherit a subset. 
Inheritance analysis is provided in Sec.~\ref{sec:inheritance_ablation} of the supplement.

\section{Experiments}

\begin{table*}[t]
  \centering
  \caption{Quantitative comparison on three datasets (\emph{1K} setting). Time is measured in minutes. $N_{\mathrm{GS}}$ denotes the number of Gaussians (in millions). The top-3 results in each column are highlighted with \colorbox{tabfirst}{\textbf{best}}, \colorbox{tabsecond}{second-best}, and \colorbox{tabthird}{third-best}.}

  \label{tab:main_quant}
  \small
  \setlength{\tabcolsep}{0.5pt}

  \resizebox{\textwidth}{!}{%
  \begin{tabular}{lcccccc cccccc cccccc}
    \toprule
    \multirow{2}{*}{Method}
    & \multicolumn{6}{c}{\textbf{Mip-NeRF 360}}
    & \multicolumn{6}{c}{\textbf{Deep Blending}}
    & \multicolumn{6}{c}{\textbf{Tanks \& Temples}} \\
    \cmidrule(lr){2-7}\cmidrule(lr){8-13}\cmidrule(lr){14-19}
    & Time$\downarrow$ & PSNR$\uparrow$ & SSIM$\uparrow$ & LPIPS$\downarrow$ & $N_{\mathrm{GS}}$ (M)$\downarrow$ & FPS$\uparrow$
    & Time$\downarrow$ & PSNR$\uparrow$ & SSIM$\uparrow$ & LPIPS$\downarrow$ & $N_{\mathrm{GS}}$ (M)$\downarrow$ & FPS$\uparrow$
    & Time$\downarrow$ & PSNR$\uparrow$ & SSIM$\uparrow$ & LPIPS$\downarrow$ & $N_{\mathrm{GS}}$ (M)$\downarrow$ & FPS$\uparrow$ \\
    \midrule
    Original 3DGS    & 14.69 & 27.58 & \third{0.814} & \third{0.220} & 2.66                  & 116 & 12.20 & 29.83 & \third{0.907} & \best{0.238} & 2.47                  & 156  & 9.47 & 23.82 & \third{0.853} & \second{0.169} & 1.58 & 157 \\
    Opti3DGS & 21.07 & 27.44 & 0.807 & 0.241 & 2.23 & 122 & 18.40 & 29.62 & 0.906 & 0.246 & 2.12 & 131 & 10.23 & 23.58 & 0.843 & 0.191 & 1.23 & \third{174} \\
    Mini-Splatting   & 19.66 & 27.36 & \best{0.822} & \best{0.216} & 0.53                    & \second{172} & 14.15 & \third{30.01} & \second{0.908} & 0.253 & 0.35                & \third{231}  & 10.55 & 23.28 & 0.835 & 0.203 & 0.20 & \best{250} \\
    Speedy-Splat     & 16.05 & 26.92 & 0.782 & 0.294 & 0.30                                  & \third{145} & 12.90 & 29.67 & 0.904 & 0.267 & 0.25                                 & \best{311}  & 8.88 & 23.42 & 0.820 & 0.240 & 0.18 & 169 \\
    Taming-3DGS      & 8.88 & \second{27.73} & 0.810 & 0.228 & 2.29                          & 97 & 6.84 & 29.29 & 0.899 & 0.246 & 2.23                                  & 123  & 6.33 & \best{24.36} & \best{0.860} & \best{0.165} & 1.49 & 142 \\
    DashGaussian     & 5.25 & 27.40 & 0.797 & 0.249 & 2.00                                   & 93 & 4.18 & 29.60 & 0.905 & 0.256 & 1.64                                  & 126  & 4.22 & \third{24.14} & 0.850 & 0.185 & 1.15 & 148\\
    FastGS           & \second{4.82} & \best{27.96} & \second{0.820} & \second{0.216} & 1.17 & \best{198}  & \second{3.34} & \best{30.24} & \best{0.911} & \second{0.240} & 0.65 & \second{237} & \second{3.44} & \second{24.34} & \second{0.858} & \third{0.173} & 0.53 & \second{208} \\
    \midrule
    Ours(Taming) & \third{5.13} & 27.39 & 0.797 & 0.245 & 1.13/1.47 & 96 & \third{3.61} & 29.38 & 0.900 & \third{0.243} & 1.02/1.20 & 127 & \third{3.91} & 23.93 & 0.841 & 0.197 & 0.57/0.88 & 110 \\
    Ours(FastGS) & \best{3.96} & \third{27.67} & 0.809 & 0.233 & 0.71/0.87 & 122 & \best{3.04} & \second{30.16} & 0.907 & 0.257 & 0.37/0.29 & 173 & \best{3.13} & 24.03 & 0.842 & 0.205 & 0.24/0.36 & 124 \\
    \bottomrule
  \end{tabular}%
  }
\end{table*}

\begin{figure*}[t]
    \centering
    \includegraphics[width=\textwidth]{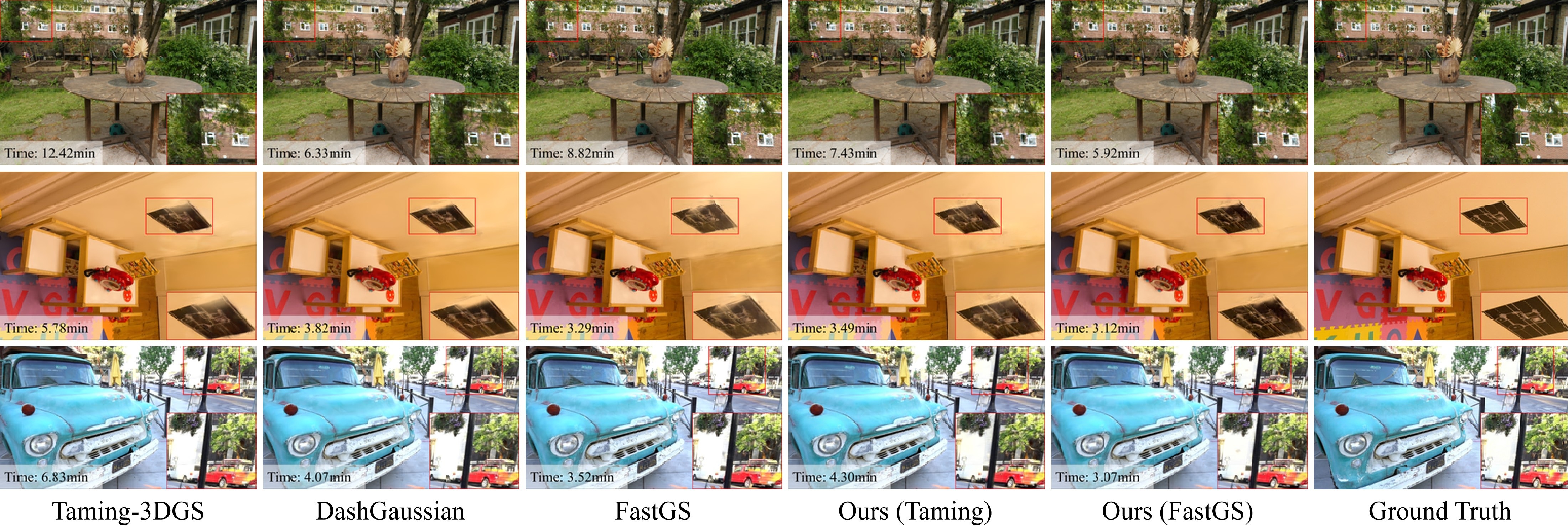}
    \vspace{-5mm}
    \caption{Qualitative results on the ``garden'' scene (Mip-NeRF 360), the ``playroom'' scene (Deep Blending), and the ``truck'' scene (Tanks \& Temples) under the \emph{1K} setting. For space constraints, we show only the top five methods by training time. Complete results and additional qualitative comparisons for the \emph{2K}/\emph{4K} settings are provided in Sec.~\ref{Scene-wise Qualitative Results} of the supplementary material.}
    \vspace{-5mm}
    \label{fig:Results of 1K}
\end{figure*}

\subsection{Experimental Setup}
\subsubsection{Datasets and Metrics.}
Following vanilla 3DGS and prior work, we evaluate our method on three real-world benchmarks: Mip-NeRF 360 \citep{barron2022mip}, Deep Blending \cite{hedman2018deep}, and Tanks\&Temples \citep{knapitsch2017tanks}. For Mip-NeRF 360, prior methods typically resize images to width 1600, which we refer to as the \emph{1K} setting. We also evaluate on full-resolution images (average width 4185), referred to as the \emph{4K} setting. We also include an intermediate \emph{2K} setting by resizing each scene to the midpoint between width 1600 and its full resolution (average width 2893).

For NVS quality, we report the PSNR (peak signal-to-noise ratio), SSIM \citep{wang2004image}, and LPIPS \citep{zhang2018unreasonable} on each dataset. To measure optimization efficiency, we report average training time and the number of primitives $N_{\mathrm{GS}}$, with stage-wise counts for our method. The reported training time measures iterative optimization for all methods. A detailed runtime breakdown and timing boundary are provided in Sec.~\ref{sec:runtime_breakdown} of the supplement.

\subsubsection{Implementation Details.} We implement our method on top of two backbones, Taming-3DGS \citep{mallick2024taming} and FastGS \citep{ren2025fastgs}. Taming-3DGS is a widely used and competitive backbone, while FastGS is a recent state-of-the-art baseline. Unless otherwise stated, all methods are run with their default settings for 30K training iterations using Adam \citep{kingma2014adam}. Experiments in the \emph{1K} setting are conducted on an NVIDIA L40 GPU, while \emph{2K} and \emph{4K} experiments are run on an NVIDIA Pro6000 GPU to avoid out-of-memory failures for some methods. We use a two-level Laplacian decomposition with an iteration split of 10K/20K. Sec.~\ref{Analysis of the number of Laplacian Levels} analyzes levels and iteration allocation.

\subsection{Comparison with Fast Optimization Methods}

\subsubsection{Baselines.} We compare against recent fast-training methods, including vanilla 3DGS \citep{kerbl3Dgaussians}, Opti3DGS \citep{farooq2025optimized}, Mini-Splatting \citep{fang2024mini}, Speedy-Splat \citep{HansonSpeedy}, Taming-3DGS \citep{mallick2024taming}, DashGaussian \citep{chen2025dashgaussian}, and FastGS \citep{ren2025fastgs}. These baselines represent different directions for accelerating 3DGS training. Taming-3DGS denotes the variant with the efficient backward implementation and the Sparse Adam optimizer. FastGS uses the “Big” setting from the original paper for a fairer comparison.
EfficientGS~\citep{liu2025efficientgs} is additionally included for high-resolution comparisons.

\subsubsection{Comparison with Other Methods on \emph{1K} setting.} Tab.~\ref{tab:main_quant} reports quantitative comparisons on three datasets under the \emph{1K} setting. Our Laplacian framework improves training efficiency on both Taming-3DGS and FastGS backbones while largely preserving reconstruction quality. On Mip-NeRF 360, Ours(Taming) reduces training time from 8.88 to 5.13 minutes, yielding a $1.73\times$ speedup with only a 0.34 dB PSNR drop. On the FastGS backbone, Ours(FastGS) further reduces time from 4.82 to 3.96 minutes, with a 0.29 dB PSNR drop. Changes in SSIM and LPIPS remain within a comparable range. On Deep Blending, Ours(Taming) is both faster and improves over the Taming baseline in quality metrics, while Ours(FastGS) also accelerates training with only minor metric degradation. Tanks \& Temples shows a similar trend. Our method achieves the fastest training time across these benchmarks. We note that slight metric degradation can occur in some cases, which we attribute to additional approximation error introduced by the hierarchical rendering and composition.

Fig.~\ref{fig:Results of 1K} presents qualitative comparisons. Despite being faster, our method preserves comparable quality and in some cases produces better renderings than prior methods. For example, in the second row, the zoomed-in view shows clearer edges and finer details in our results. In addition to static comparisons, we provide trajectory videos in the supplement to inspect temporal stability under continuous viewpoint changes. The videos demonstrate stable rendering, with no flickering artifacts introduced by the image-domain Laplacian composition. An edge-preservation analysis is also provided in Sec.~\ref{sec:boundary_edge_analysis} of the supplement.
Additional scene-wise results are provided in Sec.~\ref{Scene-wise Quantitative Results} and Sec.~\ref{Scene-wise Qualitative Results}. 

Tab.~\ref{tab:main_quant} also reports the number of Gaussian primitives $N_{\mathrm{GS}}$. For our method, we provide the stage-wise counts for each level. Compared to the original backbones, our training is split into two smaller Gaussian fields, which reduces GPU memory requirements and is better suited to resource-constrained settings. We report the peak Gaussian count and peak GPU memory in Sec.~\ref{Analysis of peak Gaussian count and peak GPU memory} of the supplement. 
The table further includes inference FPS. Since the final image is reconstructed from separately rendered frequency fields, the overall FPS can decrease despite the smaller size of each field. All reported FPS values for our method are measured using sequential rendering of the frequency fields. We include this metric to make the inference trade-off explicit.

\begin{figure}[t]
    \centering
    \includegraphics[width=1.0\linewidth]{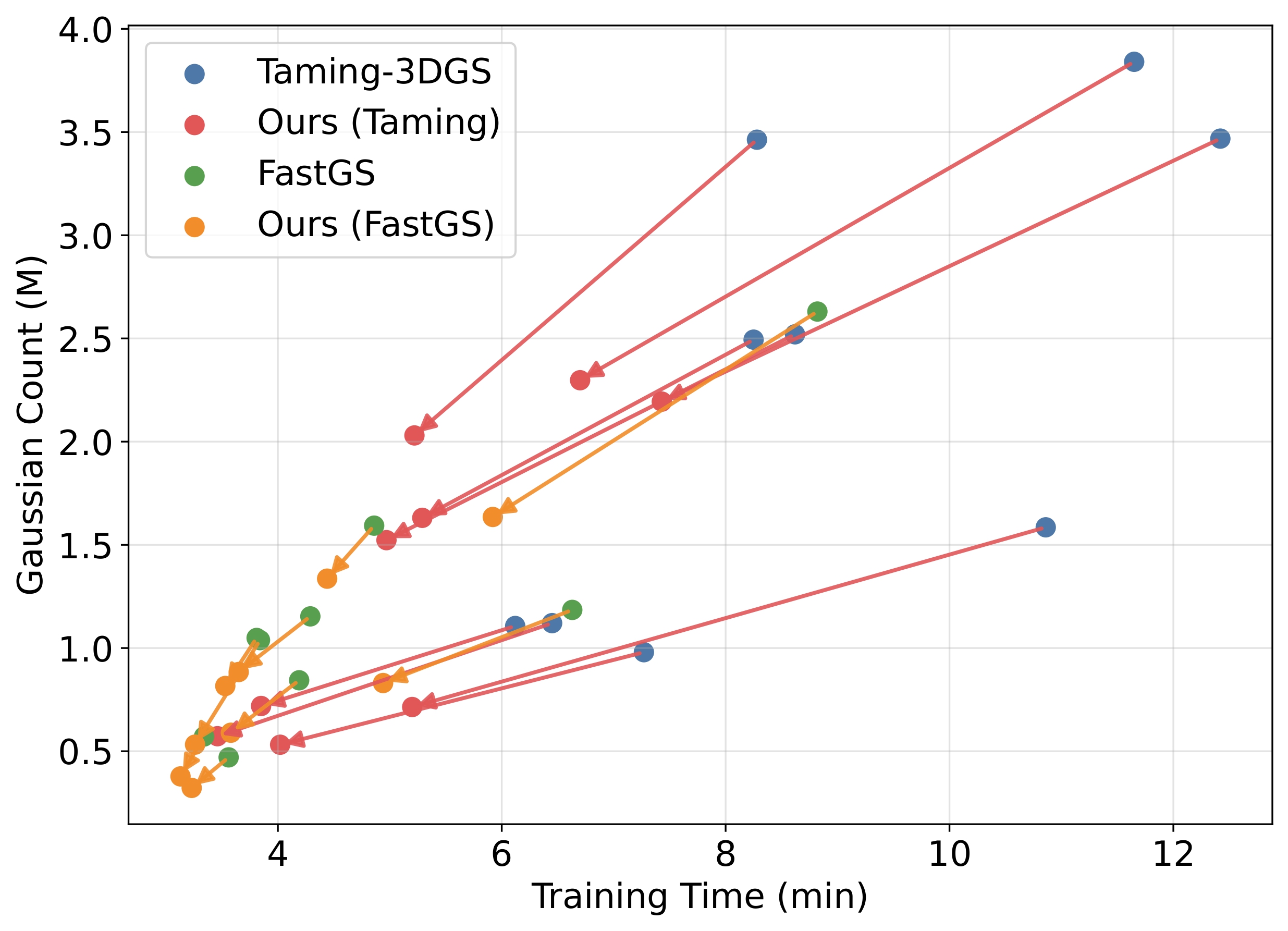}
    \vspace{-6mm}
    \caption{Training Time vs $N_{\mathrm{GS}}$ (paired before→after our method) on the Mip-NeRF 360 dataset. For our method, the $N_{\mathrm{GS}}$ is computed as an iteration-weighted average under the iteration split.}
    \vspace{-5mm}

    \label{fig:Training Time vs Gaussian Count}
\end{figure}

\begin{figure}[t]
    \centering
    \includegraphics[width=1.0\linewidth]{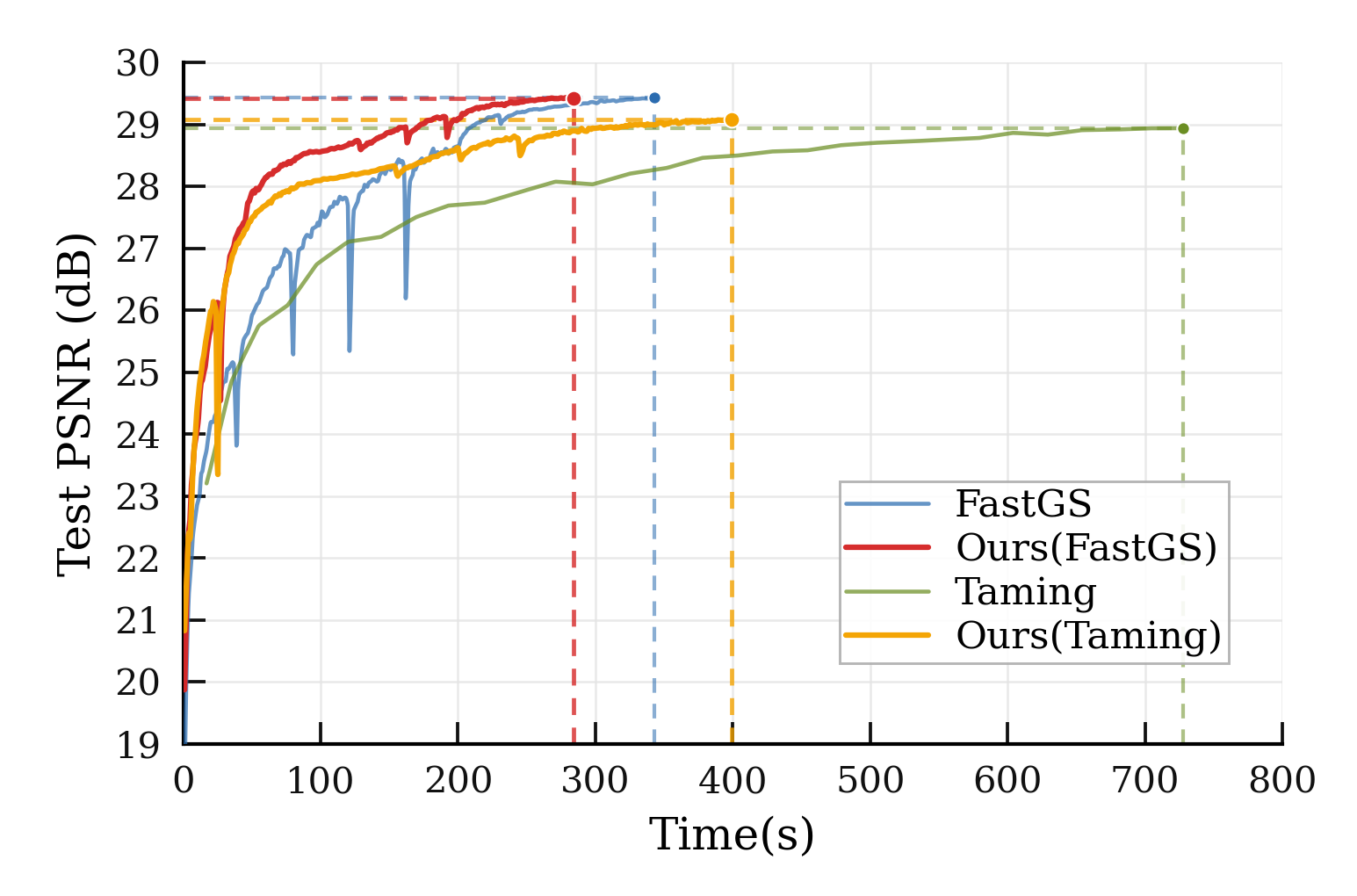}
    \vspace{-7mm}
    \caption{PSNR versus training time on the \textit{counter} scene of Mip-NeRF 360 under the 4K setting. Our method reaches comparable reconstruction quality earlier than the original backbones. The inset zooms into the early stage to highlight convergence differences.}
    \vspace{-5mm}
    \label{fig:psnr_time_4k}
\end{figure}

\subsubsection{Analysis of the number of Gaussians across training. }
Fig.~\ref{fig:number of gaussians} visualizes how the Gaussian count $N_{\mathrm{GS}}$ evolves during training. The arrows indicate the shift in $N_{\mathrm{GS}}$ when our framework is applied on top of two different backbones. We observe that in the early iterations, our curves closely follow the corresponding backbones, since the optimization behaves similarly before the densification of the first stage is completed. After densification is stopped at the base stage, the curves begin to diverge, and our method maintains a substantially lower Gaussian count thereafter. Thus, our approach yields a much lower $N_{\mathrm{GS}}$ trajectory than the original backbones, reflecting reduced effective training workload and explaining the efficiency gains observed in our experiments.

\subsubsection{Analysis of training time vs. Gaussian count.}
Fig.~\ref{fig:Training Time vs Gaussian Count} shows the paired changes in training time and average Gaussian count when applying our method on top of Taming-3DGS and FastGS. Each arrow connects a baseline run to the corresponding result with our method. 
Together with Tab.~\ref{tab:1K mip-nerf 360 scene-wise results} of the supplement, we observe that the speedup depends on scene complexity, and scenes with more Gaussians tend to obtain larger time reductions. For example, under the Taming backbone, the \textit{bicycle} scene is accelerated from 11.65 minutes to 6.70 minutes. Under the FastGS backbone, the \textit{garden} scene is accelerated from 8.82 minutes to 5.92 minutes.
When the average Gaussian count becomes sufficiently small, further reducing primitives yields a less noticeable speedup, suggesting diminishing returns for acceleration strategies that rely primarily on controlling Gaussian count. This points to an interesting direction for future work on reducing fixed rendering and optimization overheads in low-primitive regimes.

\subsubsection{Comparisons on high-resolution settings (\emph{2K} and \emph{4K})}
Tabs~\ref{tab:2k_results} and~\ref{tab:4k_results} report quantitative comparisons on the Mip-NeRF 360 dataset under the \emph{2K} and \emph{4K} settings. The overall trend is consistent with the \emph{1K} results, while the efficiency advantage of our method becomes more pronounced at higher resolutions and the quality gaps become negligible. This is expected because the rendering workload in 3DGS grows rapidly with image resolution due to heavier Gaussian--tile interactions. In particular, the number of pixel–tile evaluations increases approximately quadratically with resolution, leading to significantly higher computation and memory pressure at \emph{2K}/\emph{4K}. By reducing the effective training workload through our frequency hierarchy, our approach scales more favorably to high-resolution settings, achieving faster optimization while maintaining comparable reconstruction quality. 
The relative inference FPS gap becomes smaller at higher resolutions, especially under the 4K setting. More results can be found in the supplement.

Fig.~\ref{fig:psnr_time_4k} plots PSNR against training time under the \emph{4K} setting. Our method reaches comparable quality earlier on both backbones. These results show that the proposed framework improves not only the final training time, but also the convergence efficiency throughout optimization. Additional convergence results are provided in Sec.~\ref{sec:additional_psnr_time} of the supplement.

\begin{table}[t]
  \centering
  \caption{Mip-NeRF 360 results under the \emph{2K} setting. }
  \label{tab:2k_results}
  \small
  \setlength{\tabcolsep}{3pt}

  \resizebox{\linewidth}{!}{%
  \begin{tabular}{lcccccc}
  \toprule
    Method & Time$\downarrow$ & PSNR$\uparrow$ & SSIM$\uparrow$ & LPIPS$\downarrow$ & $N_{\mathrm{GS}}$ (M)$\downarrow$   & FPS$\uparrow$ \\
    \midrule
      Taming-3DGS & 10.37 & \third{27.14} & 0.787 & 0.313 & 1.93 & 62\\
    DashGaussian & \second{5.66} & 26.76 & 0.774 & 0.332 & 1.70 & 60\\
    FastGS & 6.18 & \best{27.53} & \best{0.804} & \best{0.282} & 1.67 & \best{121}\\
    EfficientGS & 33.35 & 27.07 & \second{0.803} & \second{0.283} & 1.26 & \third{76}\\
    Ours(Taming) & \third{6.17} & 26.87 & 0.776 & 0.325 & 1.10/1.04 & 71\\
    Ours(FastGS) & \best{4.82} & \second{27.38} & \third{0.794} & \third{0.300} & 1.22/0.70 & \second{107}\\
    \bottomrule
  \end{tabular}%
  }
\end{table}

\begin{table}[t]
  \centering
  \vspace{-2mm}
  \caption{Mip-NeRF 360 results under the \emph{4K} setting.}
  \label{tab:4k_results}
  \small
  \setlength{\tabcolsep}{3pt}

  \resizebox{\linewidth}{!}{%
  \begin{tabular}{lcccccc}
    \toprule
    Method & Time$\downarrow$ & PSNR$\uparrow$ & SSIM$\uparrow$ & LPIPS$\downarrow$ & $N_{\mathrm{GS}}$ (M)$\downarrow$ & FPS$\uparrow$\\
    \midrule
    Taming-3DGS & 17.37 & 26.81 & 0.795 & 0.361 & 1.57 & 45\\
    DashGaussian & \second{8.37} & 26.45 & 0.786 & 0.374 & 1.43 & 40\\
    FastGS & \third{8.74} & \best{27.37} & \best{0.812} & \second{0.327} & 1.73 & \best{105}\\
    EfficientGS & 68.07 & \third{26.96} & \second{0.812} & \best{0.327} & 1.24 & \third{60}\\
    Ours(Taming) & 9.96 & 26.70 & 0.790 & 0.366 & 1.00/0.79 & 48\\
    Ours(FastGS) & \best{6.57} & \second{27.22} & \third{0.807} & \third{0.337} & 1.47/0.53 & \second{95}\\
    \bottomrule
  \end{tabular}%
  }
  \vspace{-2mm}
\end{table}

\subsection{Ablation study}
\label{Ablation study}

\subsubsection{Iteration allocation.}
Tab.~\ref{tab:iter_allocation} evaluates different iteration splits between the two stages. It shows that performance is relatively insensitive to this split. This is consistent with our loss design, where the remaining reconstruction error from the previous stage is effectively delegated to the next stage. Thus, moderate reallocations of iterations mainly shift where the error is corrected rather than fundamentally changing the final quality. In our experiments, we adopt (10000, 20000) as the default since it achieves the fastest training time with comparable metrics.

\subsubsection{Analysis of the number of Laplacian levels.} 
\label{Analysis of the number of Laplacian Levels}
We investigate the effect of the number of Laplacian levels $L$ under both the \emph{1K} and \emph{4K} settings, as summarized in Tab.~\ref{tab:the number of Laplacian levels}. Under the 1K setting, a two-level hierarchy achieves the best balance between training efficiency and reconstruction quality. While increasing the number of levels leads to faster training, it consistently degrades reconstruction quality. We attribute this degradation to the accumulation of structural misalignment errors across multiple frequency bands when using deeper hierarchies. This result is consistent with prior studies~\citep{zhu2025large} on 2D Gaussian splatting for images~\citep{zhang2024gaussianimage}, which also report a two-level design as the most effective configuration.

Under the \emph{4K} setting, the quality degradation caused by deeper hierarchies becomes less pronounced, while the training speedup is significantly amplified. In particular, using four levels reduces the peak number of Gaussians from $1.74$M (2 levels) to $0.49$M, resulting in substantial savings in both memory usage and training time. These results indicate that our method benefits more evidently from deeper Laplacian hierarchies at higher resolutions, highlighting its scalability to high-resolution reconstruction. 

Overall, these results suggest that shallow hierarchies are often the most effective choice in our current setting. Although deeper hierarchies can further reduce memory usage and training time, they also tend to introduce greater quality degradation. This degradation becomes less pronounced at higher resolutions. Additional results and analyses can be found in Sec.~\ref{Levels and Iterations} of the supplement.

\begin{table}[t]
    \caption{Ablation study over the iteration allocation between two stages on the Mip-NeRF 360 dataset. }
    \label{tab:iter_allocation}
    \small
    \setlength{\tabcolsep}{4pt}
    \renewcommand{\arraystretch}{1.05}
    \resizebox{\linewidth}{!}{%
    \begin{tabular}{lcccccc}
        \toprule
        Iteration Split
        & Time$\downarrow$ & PSNR$\uparrow$ & SSIM$\uparrow$ & LPIPS$\downarrow$ & $N_{\mathrm{GS}}(M)\!\downarrow$ & FPS$\uparrow$ \\
        \midrule
        (10000, 20000) & \textbf{3.96} & 27.67 & 0.809 & 0.233 & 0.71/0.87 & 122 \\
        (15000, 15000) & 4.10 & 27.79 & 0.811 & \textbf{0.233} & 0.79/0.83 & \textbf{138} \\
        (20000, 10000) & 4.02 & \textbf{27.87} & \textbf{0.813} & 0.234 & 0.83/0.79 & 136 \\
        \bottomrule
    \end{tabular}
    }
    \vspace{-1mm}
\end{table}

\begin{table}[t!]
    \centering
\caption{Ablation studies on the number of Laplacian levels on the Mip-NeRF 360 dataset (\emph{1K} and \emph{4K} settings). For better convergence, the \emph{4K} setting here uses 45{,}000 iterations.}
    \label{tab:the number of Laplacian levels}
    \small
    \setlength{\tabcolsep}{2pt}
    \renewcommand{\arraystretch}{1.05}
    \resizebox{1.0\linewidth}{!}{%
    \begin{tabular}{cccccccc}
        \toprule
        & $L$ & Iteration Split & Time$\downarrow$ & PSNR$\uparrow$ & SSIM$\uparrow$ & LPIPS$\downarrow$ & $N_{\mathrm{GS}}(M)\!\downarrow$ \\
        \midrule
        \multirow{3}{*}{\emph{1K}}
        & 2 & (10000, 20000)
        & 3.96 & \textbf{27.67} & \textbf{0.809} & \textbf{0.233} & 0.71/0.87  \\
        & 3 & (10000, 5000, 15000)
        & 3.59 & 27.37 & 0.796 & 0.255 & 0.31/0.39/0.82  \\
        & 4 & (10000,5000,5000,10000)
        & \textbf{3.31} & 26.90 & 0.781 & 0.274 & 0.13/0.18/0.33/0.71  \\
        \midrule
        \multirow{3}{*}{\emph{4K}}
        & 2 &  (15000, 30000)
        & 10.61 & \textbf{27.39} & \textbf{0.810} & \textbf{0.331} & 1.74/0.54 \\
        & 3 &  (15000, 7500, 22500)
        & 8.65 & 27.27 & 0.803 & 0.353 & 1.07/0.63/0.38 \\
        & 4 &  (15000, 7500, 7500, 15000)
        & \textbf{7.54} & 26.97 & 0.795 & 0.370 & 0.49/0.46/0.47/0.35 \\
        \bottomrule
    \end{tabular}
    }
    \vspace{-2mm}
\end{table}

\subsubsection{Loss design.}
We study different supervision designs for the residual levels.
As shown in line 19 of Algorithm~\ref{alg:lap_hierarchy}, our default design adopts \emph{reconstructed-image supervision}, which optimizes $\mathcal{L}_{\text{rec}}^\ell = \mathcal{L}_{\mathrm{photo}}(\hat I_v^\ell, I_v^\ell)$.
Alternatively, the loss can be applied directly to the rendered residual output $\hat R_v^\ell$. We consider two such variants. 
\emph{Online residual supervision} compares the rendered residual with an online residual target computed from the previous prediction:
\[
\tilde R_{v,\mathrm{online}}^\ell
=
I_v^\ell - \mathrm{Up}(\hat I_v^{\ell+1}),
\qquad
\mathcal{L}_{\text{online-res}}^\ell
=
\mathcal{L}_{\mathrm{photo}}(\hat R_v^\ell,\tilde R_{v,\mathrm{online}}^\ell).
\]
\emph{GT Laplacian supervision} directly compares the rendered residual with the fixed ground-truth Laplacian residual $R_v^\ell$ defined in Eq.~\ref{eq:pyr_and_res}:
\[
\mathcal{L}_{\text{gt-lap}}^\ell
=
\mathcal{L}_{\mathrm{photo}}(\hat R_v^\ell, R_v^\ell).
\]

The results are shown in Tab.~\ref{tab:loss_design}. They show that \emph{reconstructed-image supervision} yields the best overall performance. Both residual-supervision variants underperform our default design, indicating that directly fitting the residual is less effective than supervising the final reconstructed image. These results support our choice of supervising the reconstructed image at each level rather than fitting the residual alone.

\begin{table}[t]
    \caption{Ablation studies over the loss design on Mip-NeRF 360.}
    \label{tab:loss_design}
    \small
    \setlength{\tabcolsep}{2pt}
    \renewcommand{\arraystretch}{1.05}
    \resizebox{\linewidth}{!}{%
    \begin{tabular}{lcccccc}
        \toprule
        Method & Time$\downarrow$ & PSNR$\uparrow$ & SSIM$\uparrow$ & LPIPS$\downarrow$ & $N_{\mathrm{GS}}(M)\!\downarrow$ & FPS$\uparrow$ \\
        \midrule
        Reconstructed-image supervision
        & \textbf{3.96} & 27.67 & \textbf{0.809} & \textbf{0.233} & 0.71/0.87 & 122 \\
        Online residual supervision
        & 4.07 & \textbf{27.70} & 0.781 & 0.250 & 0.71/0.79 & \textbf{136} \\
        GT Laplacian supervision
        & 4.11 & 27.54 & 0.804 & 0.247 & 0.71/0.80 & 136 \\
        \bottomrule
    \end{tabular}
    }
    \vspace{-1mm}
\end{table}

\section{Discussion and Limitations}
Our method is motivated by the observation that, in standard 3DGS training, coarse and low-frequency structures may converge earlier but still remain active and continue consuming optimization resources in later stages. This makes frequency-factorized training a meaningful direction for improving efficiency. However, the gain is not obtained for free. Reducing the active workload of coarse structures must be balanced against the extra cost introduced by cross-level bridging and high-frequency residual modeling. Our results also clarify both the gain and the boundary of this design space. In particular, deeper hierarchies tend to make finer residuals harder to represent efficiently with 3DGS, which may reduce the overall gain and lead to quality degradation. Our results therefore suggest that relatively shallow hierarchies already provide a better balance between speed and quality. 
To mitigate the inference overhead, the frequency fields, which are independent at inference, could be rendered in parallel. Further systems-level optimization may also improve inference efficiency, which we leave for future work.

\section{Conclusion}
We presented \textbf{Laplacian Frequency Hierarchies}, a simple and effective framework for accelerating 3DGS training by factorizing a scene into a low-frequency base field and a set of high-frequency residual fields. By combining Laplacian image decomposition with coarse-to-fine, frequency-staged training and archiving, our method reduces the number of Gaussians that remain active during optimization, improving training throughput and lowering peak GPU memory usage. The proposed framework is plug-and-play with strong 3DGS backbones, and experiments on multiple real-world benchmarks show consistent speed improvements while maintaining comparable reconstruction quality. The gains become more pronounced at higher resolutions (\emph{2K}/\emph{4K}), highlighting the practical scalability of our approach. Overall, our results suggest that frequency-factorized 3DGS is a practical direction for improving training efficiency, and we hope this work provides a useful basis for future exploration of frequency-aware 3DGS representations.

\section*{Acknowledgments}
This work was supported in part by the National Natural Science Foundation of China (NSFC) under Grant No.~62606123 and the Shenzhen Science and Technology Program under Grant No.~KJZD20240903104103005.


\bibliography{example_paper}
\bibliographystyle{icml2026}

\newpage

\appendix

\onecolumn

\begin{center}
{\Large \bfseries \papertitle}\\[0.4em]
{\Large \bfseries Supplementary Material}
\end{center}
\section{Implementation Details and Reproducibility}
\label{sec:reproducibility}

For reproducibility, we consolidate the main implementation settings here. Our framework is implemented on top of Taming-3DGS and FastGS. Unless otherwise stated, all experiments use 30K training iterations. Our default configuration uses a two-level Laplacian hierarchy with a 10K/20K iteration split. When transitioning between stages, we inherit only Gaussian positions with $\rho=1.0$ and reinitialize the remaining attributes. Each finer stage is supervised using the reconstructed image rather than directly fitting the Laplacian residual. Experiments in the 1K setting are conducted on an NVIDIA L40 GPU, while the 2K and 4K settings use an NVIDIA Pro6000 GPU. Compared methods follow their default settings unless otherwise specified in the main paper. The source code, configurations, reproduction instructions, and supplementary video are publicly available at \url{https://sorenzhang574.github.io/Laplacian-GS/}.

The supplementary video contains trajectory comparisons for several scenes under continuous viewpoint changes, which are used to inspect the temporal stability of the image-domain Laplacian composition. The results show stable rendering without visible flickering.
The video includes:
\begin{itemize}
\item a brief overview of the proposed frequency-staged training pipeline;
\item trajectory comparisons on representative scenes;
\item visualizations of the low-frequency, residual, and reconstructed components.
\end{itemize}

\section{Ablation Studies on Inheritance Design}
\label{sec:inheritance_ablation}
This section presents ablation studies on the inheritance design, including the inheritance strategy and inheritance ratio.

\subsection{Inheritance strategy: $\bm{xyz}$-only vs.\ full inheritance.}
We study how different inheritance strategies affect performance when transitioning between levels. Specifically, we compare an $xyz$-only strategy with a full inheritance strategy that retains all parameters (e.g., scale, rotation,
and optimizer states). Overall, the final reconstruction quality is relatively insensitive to this choice, which we attribute to the fact that subsequent optimization can largely correct differences in initialization. Nevertheless, the $xyz$-only strategy consistently yields slightly better results, so we adopt it as the default setting.

\subsection{Analysis of the inheritance ratio $\rho$.}
We evaluate the impact of the inheritance ratio $\rho$ on rendering quality and efficiency. As $\rho$ decreases, training becomes faster because fewer Gaussians are inherited and optimized in the next stage. However, this speedup comes at the cost of fidelity. Smaller $\rho$ leads to a clear drop in reconstruction quality, since the model has fewer primitives to represent fine details. We therefore use $\rho=1.0$ by default, which yields the best trade-off between efficiency and rendering performance.

\begin{table}[h!]
\centering
    \caption{Ablation studies on inheritance design on Mip-NeRF 360. }
    \label{tab:inheritance_ablation}
    \small
    \setlength{\tabcolsep}{4pt}
    \renewcommand{\arraystretch}{1.05}
    \begin{tabular}{lcccccc}
        \toprule
        Setting & Time$\downarrow$ & PSNR$\uparrow$ & SSIM$\uparrow$ & LPIPS$\downarrow$ & $N_{\mathrm{GS}}(M)\!\downarrow$ & FPS$\uparrow$ \\
        \midrule
        \multicolumn{7}{l}{\textbf{Inheritance strategy} (with $\rho=1.0$)} \\
        \midrule
        $xyz$-only (default) & 3.96 & \textbf{27.67} & \textbf{0.809} & \textbf{0.233} & 0.71/0.87 & 122 \\
        full inheritance   & 4.02 & 27.59 & 0.806 & 0.239 & 0.71/0.89 & 141 \\
        \midrule
        \multicolumn{7}{l}{\textbf{Inheritance ratio $\rho$} (with $xyz$-only strategy)} \\
        \midrule
        $\rho=0.5$   & 3.86 & 27.59 & 0.803 & 0.245 & 0.71/0.77 & 138 \\
        $\rho=0.25$  & 3.70 & 27.56 & 0.795 & 0.256 & 0.71/0.68 & 141 \\
        $\rho=0.125$ & \textbf{3.56} & 27.46 & 0.785 & 0.267 & 0.71/0.59 & \textbf{141} \\
        \bottomrule
    \end{tabular}
    \vspace{-1mm}
\end{table}

\section{Edge Preservation at Object Boundaries}
\label{sec:boundary_edge_analysis}

\begin{figure*}[t]
    \centering
    \includegraphics[width=\textwidth]{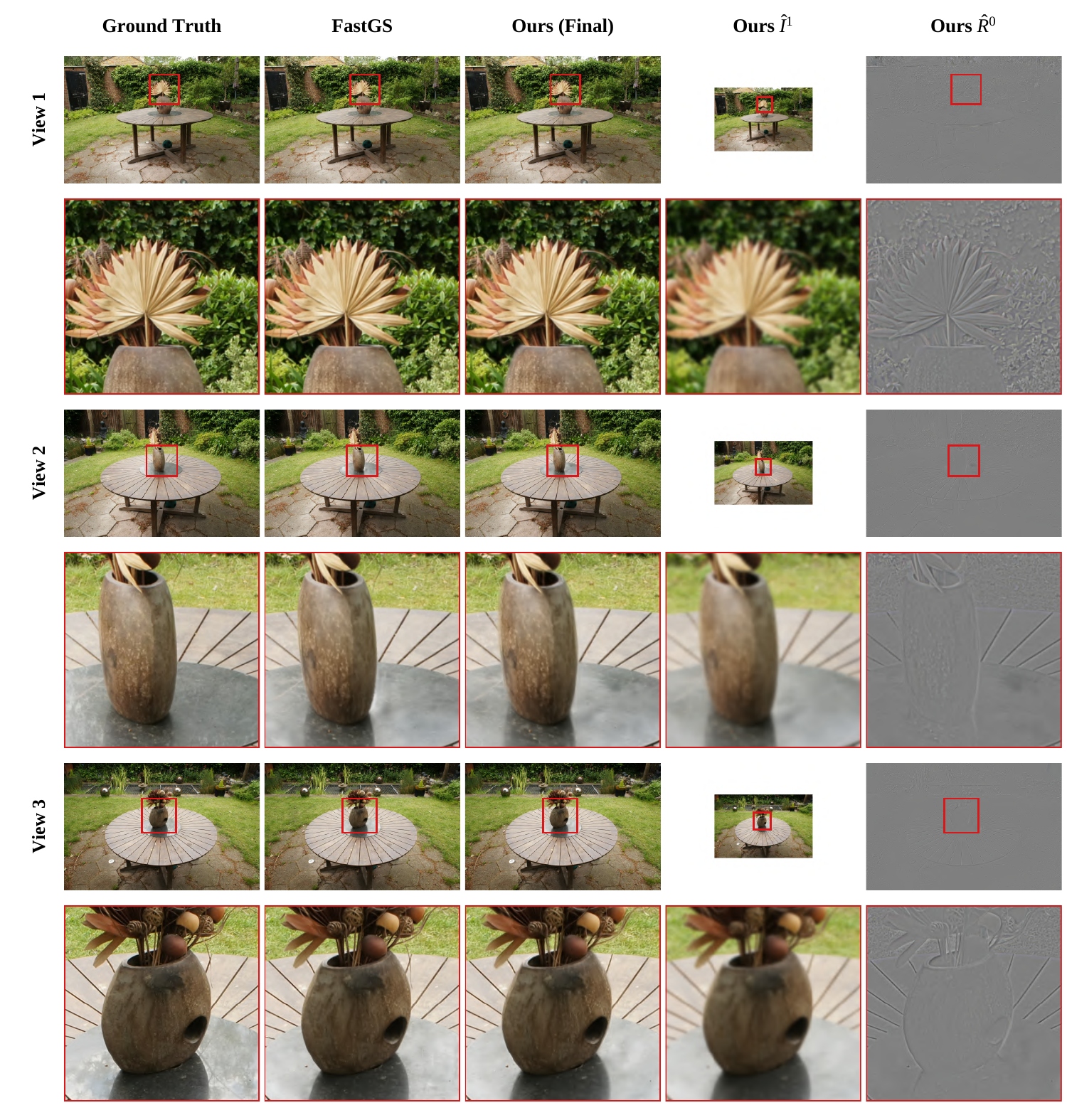}
    \caption{\textbf{Edge preservation at object boundaries on the \textit{garden} scene.}
    We show three test views with enlarged boundary regions. Although the low-frequency prediction $\hat{I}^{1}$ is smoother, the high-frequency residual $\hat{R}^{0}$ restores the missing edge details, producing final reconstructions comparable to FastGS and the GT.}
    \label{fig:boundary_edge}
\end{figure*}

We provide enlarged boundary regions to examine whether our method preserves sharp object boundaries rather than achieving stable rendering at the cost of boundary quality. As shown in Fig.~\ref{fig:boundary_edge}, although the low-frequency prediction $\hat{I}^{1}$ is smoother around object boundaries, the high-frequency residual $\hat{R}^{0}$ restores the missing edge details. The final reconstruction therefore preserves boundary sharpness and contrast comparable to those in FastGS and the GT across the three test views.

Although the image-space Laplacian is not a strictly view-invariant 3D property, Fig.~\ref{fig:boundary_edge} shows consistent boundary quality across different test views, while the video results in Sec.~\ref{sec:reproducibility} show stable reconstruction without visible flickering under continuous viewpoint changes.
In our training mechanism, each field retains the view-dependent appearance modeling of 3DGS. At each finer stage, the rendered residual is added to the cached prediction from the previous level, and the reconstructed image is directly supervised. Errors from previous levels, including viewpoint-dependent errors, can therefore be corrected by the current field. Tab.~\ref{tab:loss_design} further supports this design, as reconstructed-image supervision outperforms direct Laplacian-residual supervision.

\section{Scene-wise Quantitative Results}
\label{Scene-wise Quantitative Results}

Tab.~\ref{tab:1K mip-nerf 360 scene-wise results}, Tab.~\ref{tab:1K deep blending scene-wise results}, and Tab.~\ref{tab:1K tanks&temples scene-wise results} report scene-wise quantitative comparisons on three datasets, \mbox{Mip-NeRF 360}~\citep{barron2022mip}, Deep Blending~\citep{hedman2018deep}, and Tanks\&Temples~\citep{knapitsch2017tanks}, under the \emph{1K} setting. These results complement Tab.~\ref{tab:main_quant} in the main paper by exposing per-scene behavior. We observe that our method tends to perform better on more challenging scenes that require a larger number of Gaussians, such as \textit{garden}.

Tab.~\ref{tab:2K mip-nerf 360 scene-wise results} and Tab.~\ref{tab:4K mip-nerf 360 scene-wise results} provide scene-wise results on Mip-NeRF 360 for the \emph{2K} and \emph{4K} settings, complementing Tab.~\ref{tab:2k_results} and Tab.~\ref{tab:4k_results}, respectively. Compared to the \emph{1K} setting, our method exhibits little performance degradation at higher resolutions, and in many scenes the reconstruction metrics remain comparable or improve. Meanwhile, the acceleration gains become more pronounced, indicating that the proposed method is particularly effective when resolution increases and the training workload becomes heavier. We attribute this trend to splitting optimization into smaller Gaussian fields, which reduces the per-stage training burden and helps stabilize training at high resolutions. Overall, these results suggest that our approach not only scales well, but also tends to benefit more from higher-resolution settings.

\section{Scene-wise Qualitative Results}
\label{Scene-wise Qualitative Results}
\begin{figure*}[t]
    \centering
    \includegraphics[width=\textwidth]{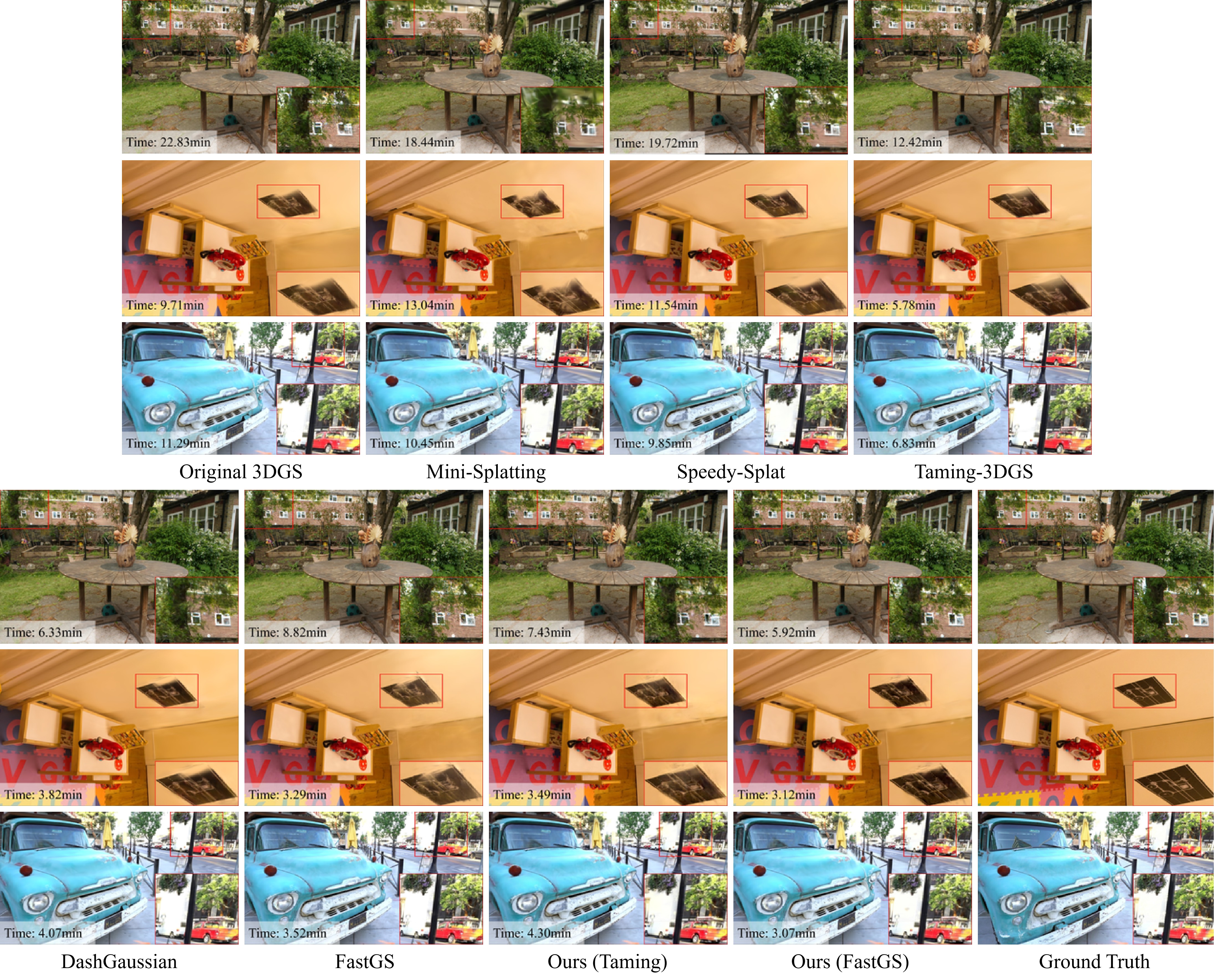}
    \caption{Qualitative results of the \emph{1K} setting. Rendering results on the "garden" scene (Mip-NeRF 360), the "playroom" scene (Deep Blending), and the "truck" scene (Tanks \& Temples).}

    \label{fig:complete results of 1K}
\end{figure*}

\begin{figure*}[t]
    \centering
    \includegraphics[width=\textwidth]{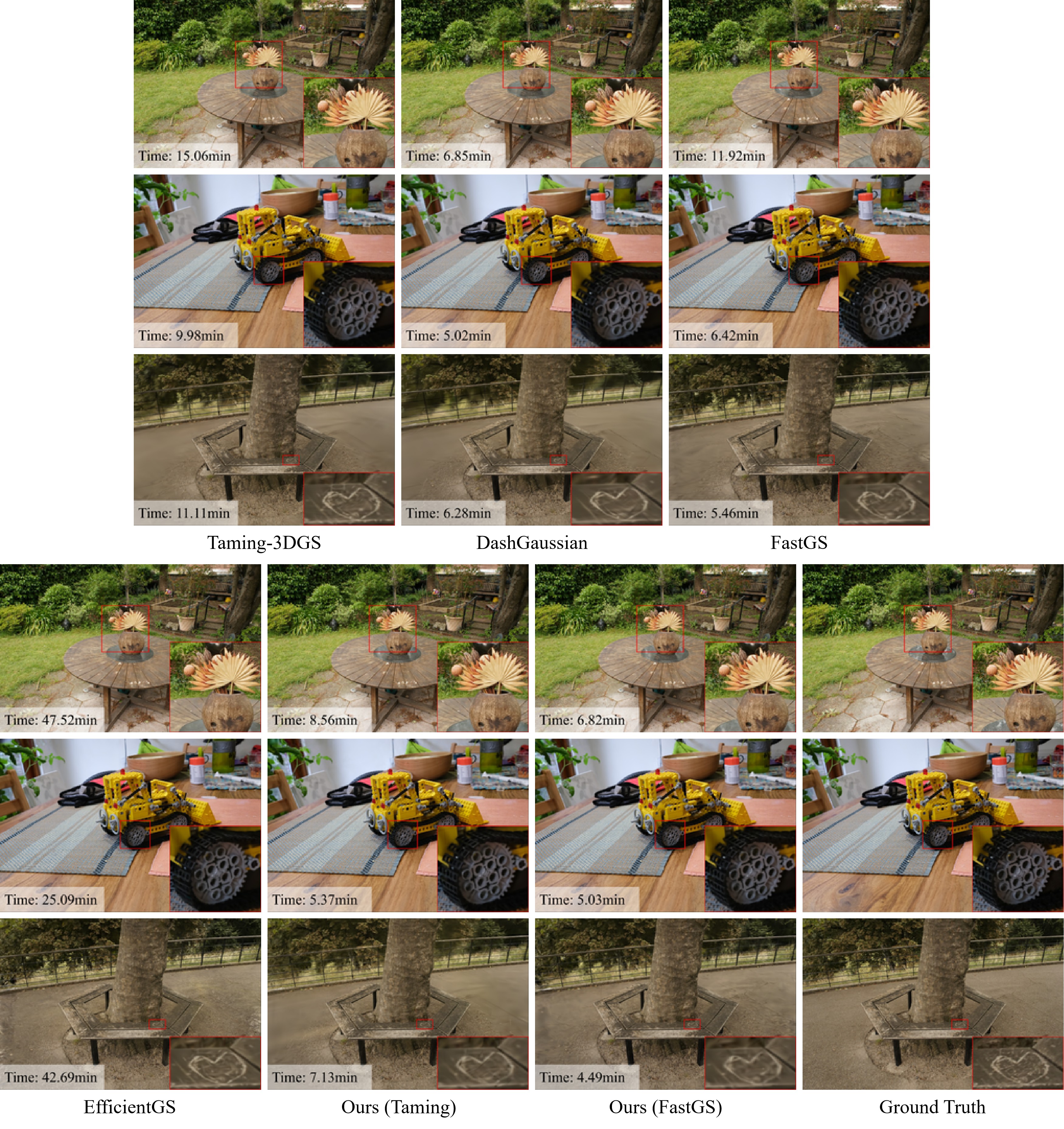}
    \caption{Qualitative results of the 2K setting. Rendering results on the "garden", "kitchen" and "treehill" scenes of Mip-NeRF 360.} 

    \label{fig:complete results of 2K}
\end{figure*}

\begin{figure*}[t]
    \centering
    \includegraphics[width=\textwidth]{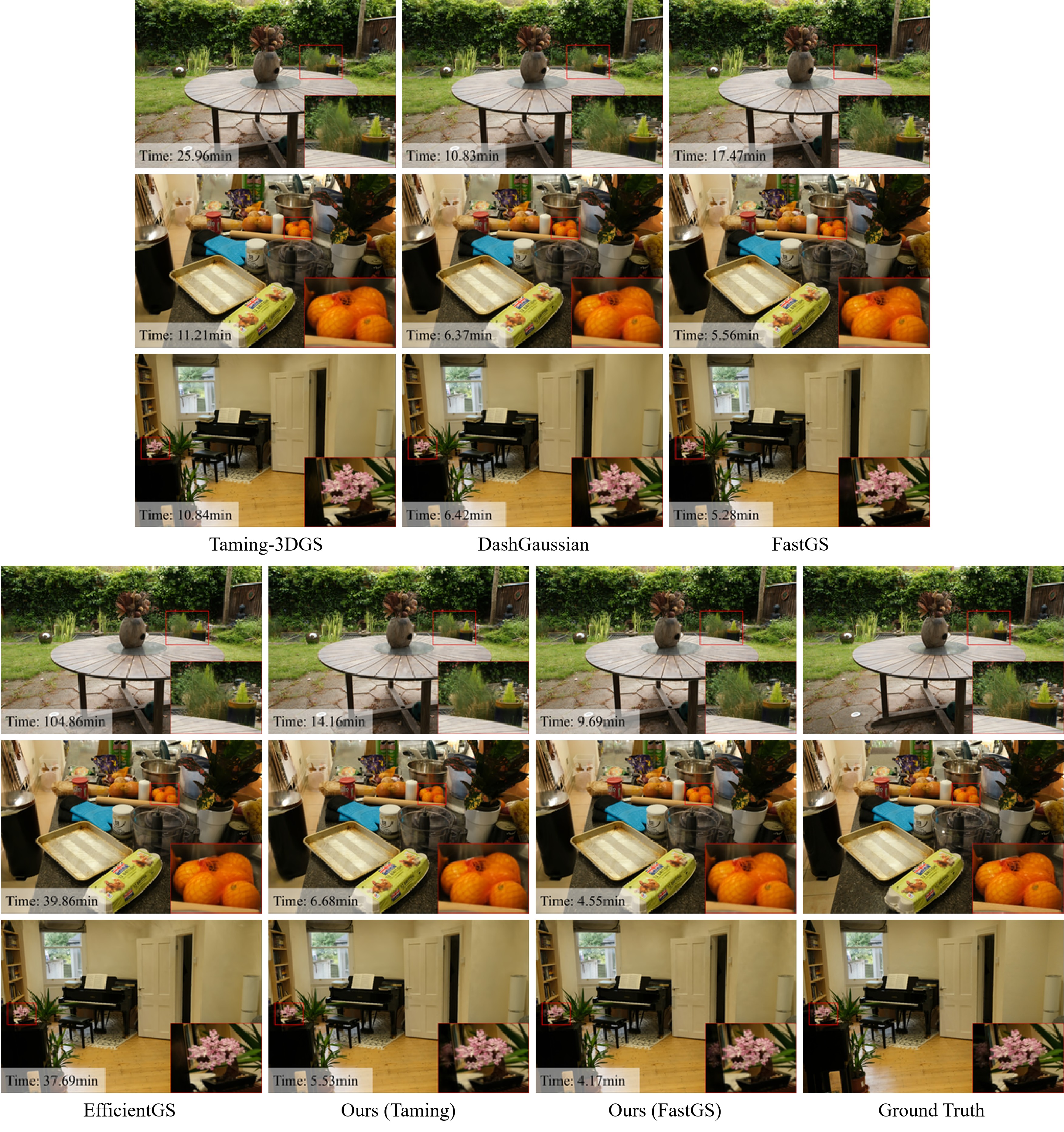}
    \caption{Qualitative results of the \emph{4K} setting. Rendering results on the "garden", "counter" and "room" scenes of Mip-NeRF 360.} 
    
    \label{fig:complete results of 4K}
\end{figure*}

Qualitative comparisons for the \emph{1K}, \emph{2K}, and \emph{4K} settings are shown in Fig.~\ref{fig:complete results of 1K}, Fig.~\ref{fig:complete results of 2K}, and Fig.~\ref{fig:complete results of 4K}, respectively. Across all settings, our method consistently maintains high visual quality. These results suggest that composing the final rendering by accumulating a sequence of components is a viable strategy and does not introduce noticeable artifacts in practice.

\section{Analysis of peak Gaussian count and peak GPU memory}
\label{Analysis of peak Gaussian count and peak GPU memory}
In Tab.~\ref{tab:avg_gs_gpu}, we report the average peak number of Gaussians and the average peak GPU memory usage during training on the Mip-NeRF 360 dataset. Note that we measure only the memory footprint of the 3DGS training pipeline and exclude memory used to store the dataset on the GPU. Overall, peak GPU memory usage is positively correlated with peak Gaussian count. Among all methods, Ours(FastGS) achieves the lowest peak Gaussian number and the lowest peak GPU memory usage.

\begin{table}[h!]
  \centering
  \caption{Average peak Gaussian count and average peak GPU memory usage during training on Mip-NeRF 360.}
  \label{tab:avg_gs_gpu}
  \small
  \setlength{\tabcolsep}{6pt}
  \renewcommand{\arraystretch}{1.05}

  \begin{tabular}{lcc}
    \toprule
    Method & Average Peak $N_{\mathrm{GS}}$ $\downarrow$ & Average Peak GPU (MB)$\downarrow$ \\
    \midrule
    Taming-3DGS   & 2{,}294{,}880 & 4{,}320 \\
    DashGaussian  & 1{,}996{,}751 & 4{,}021 \\
    FastGS        & 1{,}477{,}507 & 2{,}623 \\
    Ours(Taming)  & 1{,}621{,}565 & 3{,}144 \\
    Ours(FastGS)  & \textbf{1{,}230{,}462} & \textbf{2{,}198} \\
    \bottomrule
  \end{tabular}%
\end{table}

\section{Runtime Breakdown and Timing Boundary}
\label{sec:runtime_breakdown}

We profile a full Ours(FastGS) run of 30K iterations on the \textit{garden} scene using CUDA events. The iterative optimization takes 361.84\,s in total, corresponding to 12.061\,ms per iteration. Table~\ref{tab:runtime_breakdown} reports the runtime breakdown per iteration. Backward propagation accounts for the largest portion (56.5\%), followed by forward rendering (18.2\%) and the optimizer step (14.3\%). Composition and loss computation account for only 3.5\% of the optimization time.

\begin{table}[t]
    \centering
    \caption{Runtime breakdown of a 30K-iterations Ours(FastGS) run on the \textit{garden} scene. Percentages are measured relative to the iterative optimization time.}
    \label{tab:runtime_breakdown}
    \setlength{\tabcolsep}{5pt}
    \begin{tabular}{lcc}
        \toprule
        Component & Time (ms/iter.) & Percentage \\
        \midrule
        Forward rendering             & 2.198 & 18.2\% \\
        Composition + loss             & 0.418 & 3.5\% \\
        Backward propagation           & 6.810 & 56.5\% \\
        Densification + pruning        & 0.785 & 6.5\% \\
        Optimizer step                 & 1.722 & 14.3\% \\
        Other                          & 0.128    & 1.0\% \\
        \midrule
        Total                           & 12.061 & 100\% \\
        \bottomrule
    \end{tabular}
\end{table}

The training times reported in the main paper measure iterative optimization only. Pyramid preprocessing and archive I/O to disk are excluded. During iterative optimization, Laplacian calculation is included in the composition and loss time. The additional operations performed once consist of pyramid construction (0.018\,s), construction of the stage cache in memory, including rendering and upsampling (0.405\,s), and inheritance (0.136\,s), totaling only 0.559\,s, or 0.15\% of the iterative optimization time.

Archive I/O to disk is also excluded from the reported training times, consistent with excluding model saving and checkpoint I/O for the compared methods. For reference, the two archive writes take 16.72\,s in total.

\section{Additional analysis on Laplacian levels and training iterations}
\label{Levels and Iterations}
We provide additional ablation results on the number of Laplacian levels and training iterations in Tabs.~\ref{tab:the number of Laplacian levels, 1k} and~\ref{tab:more iter./levels results on 4K dataset}. 
Tab.~\ref{tab:the number of Laplacian levels, 1k} reports detailed results under the \emph{1K} setting. 
Consistent with the main paper, a two-level hierarchy achieves the best overall trade-off, while deeper hierarchies reduce training time but lead to a gradual degradation in reconstruction quality and rendering speed.

Tab.~\ref{tab:more iter./levels results on 4K dataset} further investigates the effect of increasing the total training iterations under the \emph{4K} setting.
We report results with both 30k and 45k iterations, where the two settings share identical split ratios across levels.
We include 45k iterations to ensure better convergence in the high-resolution setting.
As the iteration budget increases, reconstruction quality slightly improves across all level choices, while the relative trends among different hierarchies remain consistent.
In particular, deeper Laplacian hierarchies consistently reduce the peak number of Gaussians and training time, and the associated degradation in rendering quality becomes less noticeable at higher resolutions.
These results further support our observation that the proposed Laplacian decomposition is especially effective in high-resolution regimes.

\section{Additional results for PSNR versus training time}
\label{sec:additional_psnr_time}

\begin{figure}[t]
    \centering
    \begin{minipage}[t]{0.49\linewidth}
        \centering
        \includegraphics[width=\linewidth]{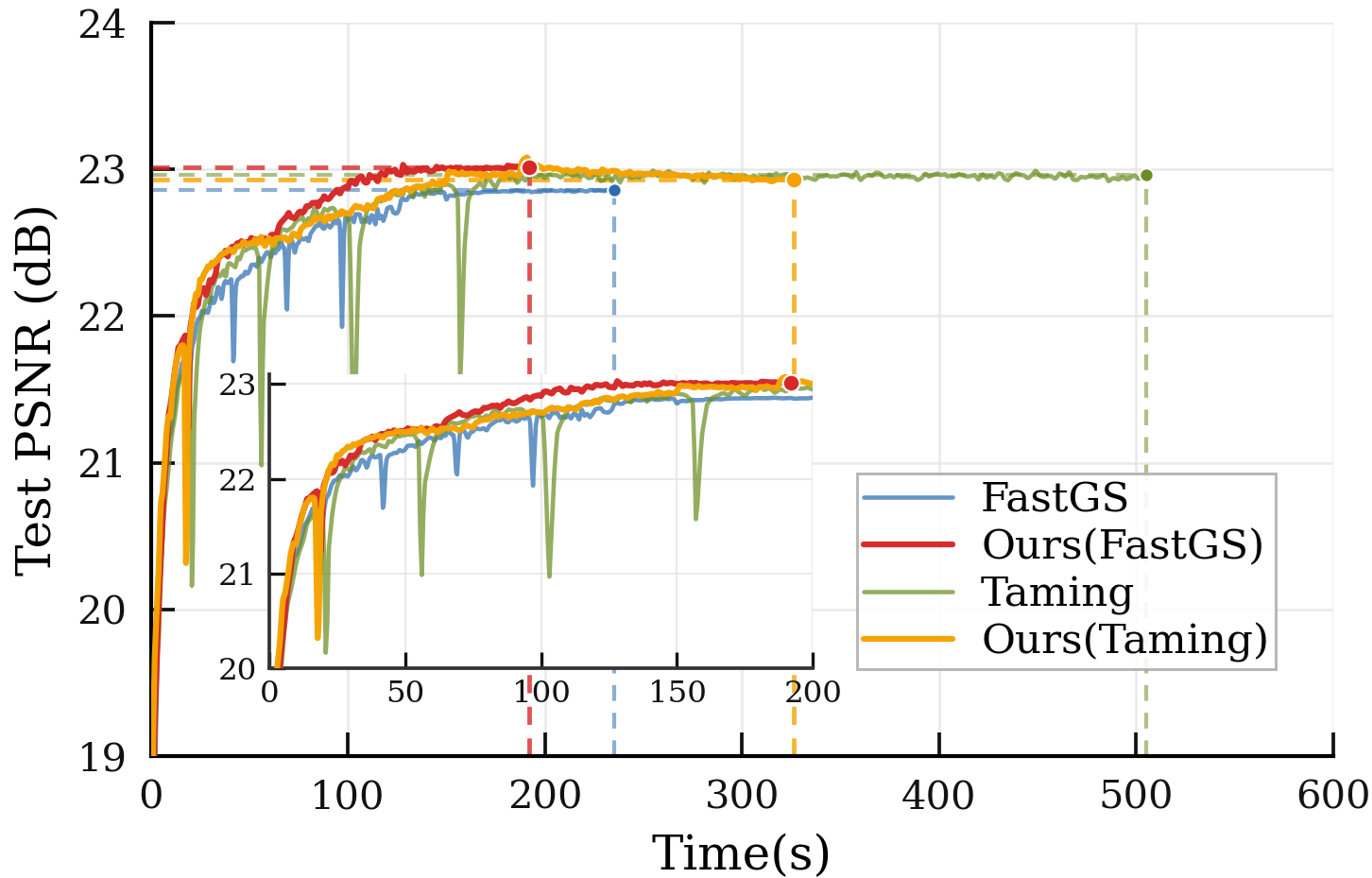}
        
        {\small (a) 1K setting (\textit{treehill})}
    \end{minipage}
    \hfill
    \begin{minipage}[t]{0.49\linewidth}
        \centering
        \includegraphics[width=\linewidth]{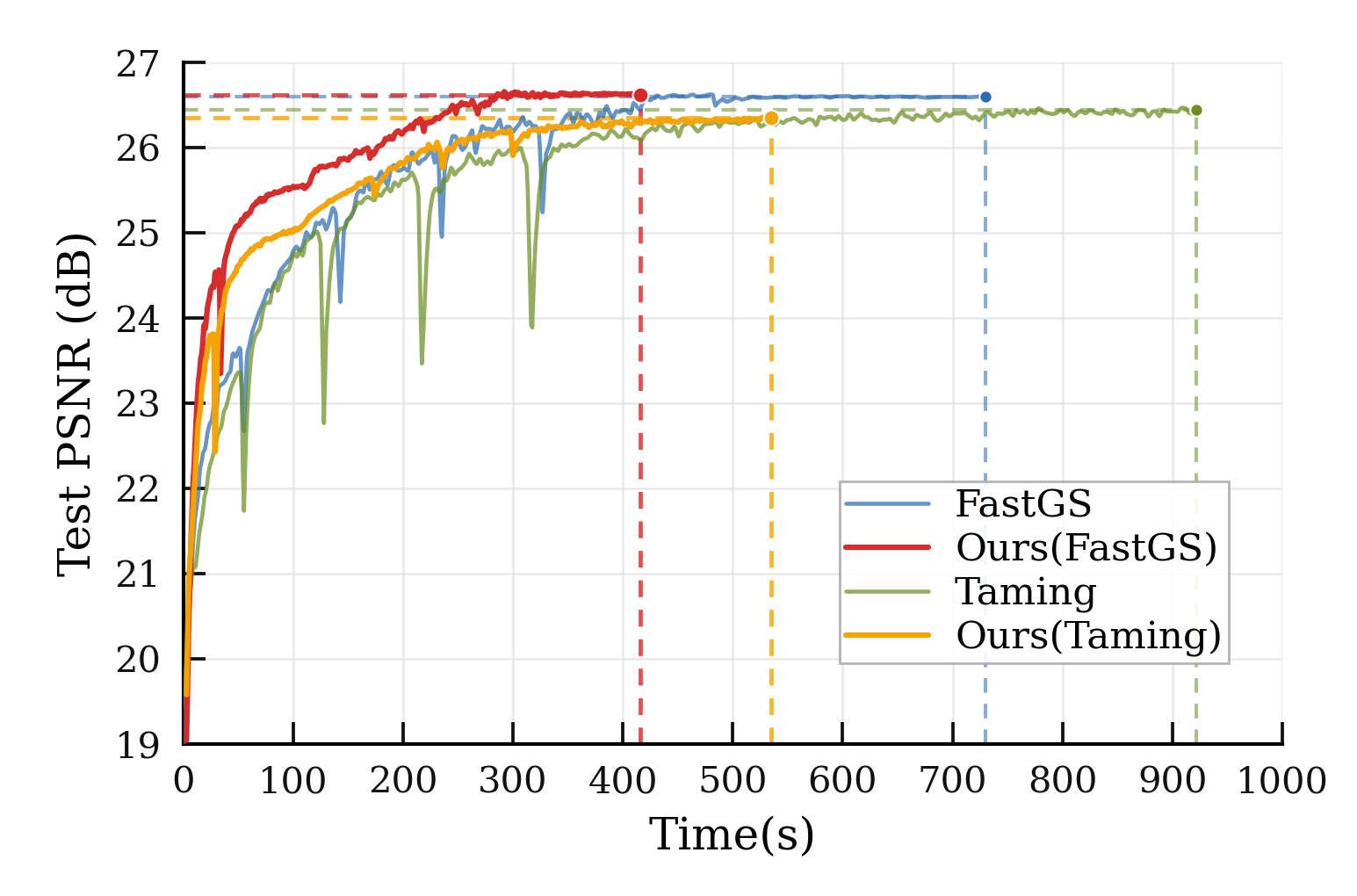}
        
        {\small (b) 2K setting (\textit{garden})}
    \end{minipage}
    \caption{PSNR versus training time under the \emph{1K} and \emph{2K} settings on Mip-NeRF 360 scenes. Similar to the \emph{4K} case in the main paper, our method reaches comparable reconstruction quality earlier than the original backbones. }
    \label{fig:psnr_time_1k_2k}
\end{figure}

In the main paper, we present the PSNR versus training time curves under the \emph{4K} setting in Fig.~\ref{fig:psnr_time_4k}. Due to space constraints, we provide the corresponding results for the \emph{1K} and \emph{2K} settings in Fig.~\ref{fig:psnr_time_1k_2k}. Similar to the \emph{4K} case, our method reaches comparable reconstruction quality earlier than the original backbones on both Taming-3DGS and FastGS. Together with the \emph{4K} result in the main paper, the advantage is more evident at higher resolutions, where the training workload is heavier and the benefit of reducing the active Gaussian workload is more pronounced. These results further support the convergence efficiency of the proposed framework throughout optimization.

\begin{table*}[t]
    \centering
    \caption{Results of the \emph{1K} setting. Scene-wise quantitative results over Mip-NeRF 360 dataset. Time is measured in minutes. $N_{\mathrm{GS}}$ denotes the number of Gaussians (in millions).}
    \label{tab:1K mip-nerf 360 scene-wise results}
    \small
    \setlength{\tabcolsep}{4pt}
    \renewcommand{\arraystretch}{1.05}
  
    \resizebox{\textwidth}{!}{ 
    \begin{tabular}{lcccccc cccccc cccccc}
        \toprule
        \multirow{2}{*}{Method}
        & \multicolumn{6}{c}{\textbf{bicycle}} & \multicolumn{6}{c}{\textbf{bonsai}} & \multicolumn{6}{c}{\textbf{counter}} \\
        \cmidrule(lr){2-7} \cmidrule(lr){8-13} \cmidrule(lr){14-19}
        & Time$\downarrow$ & PSNR$\uparrow$ & SSIM$\uparrow$ & LPIPS$\downarrow$ & $N_{\mathrm{GS}}(M)\!\downarrow$ & FPS$\uparrow$ & Time$\downarrow$ & PSNR$\uparrow$ & SSIM$\uparrow$ & LPIPS$\downarrow$ & $N_{\mathrm{GS}}(M)\!\downarrow$ & FPS$\uparrow$ & Time$\downarrow$ & PSNR$\uparrow$ & SSIM$\uparrow$ & LPIPS$\downarrow$ & $N_{\mathrm{GS}}(M)\!\downarrow$ & FPS$\uparrow$ \\
        \midrule
        Original 3DGS  
        & 23.27 & \third{25.18} & \third{0.748} & \second{0.242} & 4.78 & 79  
        & 8.02  & 32.40 & 0.947 & 0.180 & 1.07  & 123 
        & 9.10  & 29.17 & \third{0.916} & \third{0.183} & 1.05 & 104 \\
        Opti3DGS
        & 30.37 & 25.08 & 0.734 & 0.277 & 4.26 & 60
        & 15.63 & 32.03 & 0.944 & 0.186 & 0.97 & \best{199}
        & 16.35 & 29.03 & 0.910 & 0.194 & 0.79 & \best{160} \\
        Mini-Splatting 
        & 18.26 & \second{25.22} & \best{0.764} & \best{0.241} & 0.59 & \best{137} 
        & 19.27 & 31.41 & 0.944 & \third{0.176} & 0.36  & \third{174} 
        & 22.59 & 28.61 & 0.912 & \second{0.181} & 0.41 & \third{143} \\
        Speedy-Splat   
        & 19.87 & 24.72 & 0.704 & 0.333 & 0.58 & \second{130} 
        & 13.16 & 31.28 & 0.926 & 0.227 & 0.13  & \second{185} 
        & 13.81 & 28.30 & 0.877 & 0.258 & 0.10 & 128 \\
        Taming-3DGS    & 11.65 & 25.01 & 0.731 & 0.269 & 3.84 & 76  
        & 6.12  & \second{32.73} & \second{0.948} & 0.180 & 1.11 & 117 
        & 7.27  & \third{29.43} & \second{0.916} & 0.183 & 0.98 & 109 \\
        DashGaussian   
        & 6.96  & 24.91 & 0.723 & 0.284 & 3.78 & 71  
        & \third{3.87}  & 32.01 & 0.941 & 0.192 & 0.80  & 122 
        & 4.31  & 28.94 & 0.907 & 0.202 & 0.72 & 96 \\
        FastGS         
        & \second{4.86}  & \best{25.26} & \second{0.755} & \third{0.245} & 1.59 & 72  
        & 4.19  & \best{33.08} & \best{0.954} & \best{0.160} & 0.84  & 115 
        & \second{3.56}  & \best{29.60} & \best{0.918} & \best{0.177} & 0.47 & 103 \\
        Ours (Taming)  
        & \third{6.70} & 24.76 & 0.720 & 0.282 & 1.74/2.57 & 72  
        & \second{3.85} & 32.17 & 0.942 & 0.188 & 0.87/0.64 & 115 
        & \third{4.02} & 29.22 & 0.908 & 0.198 & 0.58/0.51 & 103 \\
        Ours (FastGS)  
        & \best{4.44} & 25.00 & 0.742 & 0.267 & 1.11/1.45 & \third{116} 
        & \best{3.58} & \third{32.43} & \third{0.948} & \second{0.171} & 0.74/0.52 & 117 
        & \best{3.23} & \second{29.43} & 0.913 & 0.189 & 0.39/0.29 & \second{146} \\
        \midrule
        \midrule
        \multirow{2}{*}{Method}
        & \multicolumn{6}{c}{\textbf{flowers}} & \multicolumn{6}{c}{\textbf{garden}} & \multicolumn{6}{c}{\textbf{kitchen}} \\
        \cmidrule(lr){2-7} \cmidrule(lr){8-13} \cmidrule(lr){14-19}
        & Time$\downarrow$ & PSNR$\uparrow$ & SSIM$\uparrow$ & LPIPS$\downarrow$ & $N_{\mathrm{GS}}(M)\!\downarrow$ & FPS$\uparrow$ & Time$\downarrow$ & PSNR$\uparrow$ & SSIM$\uparrow$ & LPIPS$\downarrow$ & $N_{\mathrm{GS}}(M)\!\downarrow$ & FPS$\uparrow$ & Time$\downarrow$ & PSNR$\uparrow$ & SSIM$\uparrow$ & LPIPS$\downarrow$ & $N_{\mathrm{GS}}(M)\!\downarrow$ & FPS$\uparrow$ \\
        \midrule
        Original 3DGS  
        & 14.58 & 21.38 & 0.588 & \third{0.358} & 2.85 & \third{138} 
        & 22.83 & 27.36 & \third{0.858} & \third{0.122} & 4.21 & 98  
        & 11.93 & 31.56 & \third{0.933} & \third{0.116} & 1.55 & 133 \\
        Opti3DGS
        & 21.53 & \second{21.61} & \third{0.595} & 0.359 & 2.62 & 110
        & 22.73 & 27.11 & 0.835 & 0.171 & 3.05 & 83
        & 18.50 & 31.33 & 0.928 & 0.126 & 1.00 & 153 \\
        Mini-Splatting 
        & 19.35 & 21.31 & \best{0.614} & \second{0.340} & 0.64 & \best{268} 
        & 18.44 & 26.84 & 0.840 & 0.161 & 0.67 & \best{239} 
        & 22.43 & 31.41 & 0.931 & 0.120 & 0.43 & \best{187} \\
        Speedy-Splat   
        & 16.20 & 21.17 & 0.560 & 0.418 & 0.34 & 135 
        & 19.72 & 26.74 & 0.815 & 0.213 & 0.52 & \third{134} 
        & 15.16 & 29.94 & 0.895 & 0.194 & 0.11 & \third{162} \\
        Taming-3DGS    
        & 8.62  & \third{21.52} & 0.590 & 0.361 & 2.52 & 107 
        & 12.42 & \third{27.48} & 0.854 & 0.130 & 3.47 & 72  
        & 10.86 & \third{31.82} & 0.932 & 0.117 & 1.58 & 86 \\
        DashGaussian   
        & 5.75  & 21.18 & 0.564 & 0.386 & 2.35 & 98  
        & \second{6.33}  & 27.17 & 0.836 & 0.165 & 2.53 & 74  
        & \second{5.05}  & 31.53 & 0.925 & 0.131 & 1.07 & 94   \\
        FastGS         
        & \second{4.29}  & \best{21.63} & \second{0.604} & \best{0.338} & 1.15 & \second{209} 
        & 8.82  & \best{27.61} & \best{0.865} & \best{0.110} & 2.63 & \second{163} 
        & 6.63  & \best{32.39} & \best{0.939} & \best{0.104} & 1.18 & \second{185} \\
        Ours (Taming)  
        & \third{4.97} & 21.04 & 0.547 & 0.400 & 1.17/1.70 & 102 
        & \third{7.43} & 27.35 & 0.849 & 0.143 & 1.10/2.74 & 79 
        & \third{5.20} & 31.51 & 0.927 & 0.129 & 0.81/0.67 & 87 \\
        Ours (FastGS)  
        & \best{3.65} & 21.02 & 0.556 & 0.383 & 0.74/0.95 & 134 
        & \best{5.92} & \second{27.56} & \second{0.862} & \second{0.121} & 0.93/1.99 & 86 
        & \best{4.94} & \second{32.27} & \second{0.938} & \second{0.108} & 0.82/0.84 & 123 \\
        \midrule
        \midrule
        \multirow{2}{*}{Method}
        & \multicolumn{6}{c}{\textbf{room}} & \multicolumn{6}{c}{\textbf{stump}} & \multicolumn{6}{c}{\textbf{treehill}} \\
        \cmidrule(lr){2-7} \cmidrule(lr){8-13} \cmidrule(lr){14-19}
        & Time$\downarrow$ & PSNR$\uparrow$ & SSIM$\uparrow$ & LPIPS$\downarrow$ & $N_{\mathrm{GS}}(M)\!\downarrow$ & FPS$\uparrow$ & Time$\downarrow$ & PSNR$\uparrow$ & SSIM$\uparrow$ & LPIPS$\downarrow$ & $N_{\mathrm{GS}}(M)\!\downarrow$ & FPS$\uparrow$ & Time$\downarrow$ & PSNR$\uparrow$ & SSIM$\uparrow$ & LPIPS$\downarrow$ & $N_{\mathrm{GS}}(M)\!\downarrow$ & FPS$\uparrow$ \\
        \midrule
        Original 3DGS  
        & 8.92  & \third{31.86} & 0.928 & 0.196 & 1.26 & 161 
        & 18.36 & 26.66 & \third{0.768} & \third{0.243} & 4.04 & 104 
        & 15.16 & 22.67 & \third{0.637} & \second{0.345} & 3.08 & 103 \\
        Opti3DGS
        & 17.17 & 31.50 & 0.922 & 0.212 & 1.03 & 163
        & 24.07 & 26.74 & 0.767 & 0.262 & 3.40 & 77
        & 23.28 & 22.55 & 0.625 & 0.380 & 2.91 & 91 \\
        Mini-Splatting 
        & 19.72 & 31.41 & \second{0.929} & \second{0.189} & 0.39 & \third{185} 
        & 17.77 & \best{27.35} & \best{0.806} & \best{0.215} & 0.66 & 99  
        & 19.09 & 22.68 & \best{0.656} & \best{0.326} & 0.63 & \third{117} \\
        Speedy-Splat   
        & 13.89 & 30.95 & 0.904 & 0.256 & 0.11 & \second{201} 
        & 17.17 & 26.70 & 0.765 & 0.288 & 0.47 & \second{114} 
        & 15.46 & 22.46 & 0.590 & 0.463 & 0.33 & 115 \\
        Taming-3DGS    
        & 6.45  & \second{32.04} & 0.926 & 0.201 & 1.12 & 103 
        & 8.28  & 26.53 & 0.760 & 0.259 & 3.46 & 96  
        & 8.25  & 22.99 & \second{0.637} & \third{0.354} & 2.49 & 105 \\
        DashGaussian   
        & 4.04  & 31.72 & 0.919 & 0.217 & 0.79 & 100 
        & \third{4.94}  & 26.05 & 0.733 & 0.301 & 3.06 & 87  
        & 6.04  & \best{23.08} & 0.629 & 0.368 & 2.89 & 95 \\
        FastGS         
        & \second{3.34}  & \best{32.14} & \best{0.930} & \best{0.189} & 0.57 & \best{264} 
        & \second{3.81}  & \second{27.16} & \second{0.786} & \second{0.240} & 1.05 & \best{163} 
        & \second{3.84}  & 22.81 & 0.632 & 0.377 & 1.04 & \best{207} \\
        Ours (Taming)  
        & \third{3.46} & 31.45 & 0.921 & 0.211 & 0.58/0.57 & 125 
        & 5.22 & 26.00 & 0.733 & 0.282 & 1.97/2.06 & 86  
        & \third{5.29} & \second{23.01} & 0.623 & 0.375 & 1.37/1.76 & 95 \\
        Ours (FastGS)  
        & \best{3.13} & 31.54 & \third{0.928} & \third{0.194} & 0.40/0.19 & 136 
        & \best{3.53} & \third{26.80} & 0.767 & 0.263 & 0.78/0.83 & \third{108} 
        & \best{3.26} & \third{22.99} & 0.629 & 0.400 & 0.48/0.56 & \second{136} \\
        \bottomrule
    \end{tabular}
    } 
\end{table*}

\begin{table*}[t]
    \centering
    \caption{Results of the \emph{1K} setting. Scene-wise quantitative results over Deep Blending dataset. Time is measured in minutes. $N_{\mathrm{GS}}$ denotes the number of Gaussians (in millions).}
    \label{tab:1K deep blending scene-wise results}
    \small
    \setlength{\tabcolsep}{4pt}
    \renewcommand{\arraystretch}{1.05}

    \resizebox{!}{0.9\height}{
    \begin{tabular}{lcccccc cccccc}
        \toprule
        \multirow{2}{*}{Method}
        & \multicolumn{6}{c}{\textbf{drjohnson}} 
        & \multicolumn{6}{c}{\textbf{playroom}} \\
        \cmidrule(lr){2-7} \cmidrule(lr){8-13}
        & Time$\downarrow$ & PSNR$\uparrow$ & SSIM$\uparrow$ & LPIPS$\downarrow$
        & $N_{\mathrm{GS}}(M)\downarrow$ & FPS$\uparrow$
        & Time$\downarrow$ & PSNR$\uparrow$ & SSIM$\uparrow$ & LPIPS$\downarrow$
        & $N_{\mathrm{GS}}(M)\downarrow$ & FPS$\uparrow$ \\
        \midrule

        Original 3DGS 
        & 14.69 & 29.49 & \second{0.905} & \best{0.236} & 3.12  & 141 
        & 9.71 & 30.16 & 0.909 & \third{0.241} & 1.83 & \third{170} \\
        Opti3DGS
        & 21.58 & 29.22 & 0.903 & 0.244 & 2.71 & 94
        & 15.22 & 30.02 & 0.909 & 0.247 & 1.52 & 168 \\
        Mini-Splatting 
        & 15.25 & \third{29.55} & 0.904 & 0.256  & 0.38 & \second{293} 
        & 13.04 & \third{30.47} & \third{0.912} & 0.249  & 0.32 & 169 \\
        Speedy-Splat 
        & 14.26 & 29.18 & 0.900  & 0.266 & 0.31 & \best{303} 
        & 11.54 & 30.16 & 0.907 & 0.268 & 0.19 & \best{320} \\
        Taming-3DGS 
        & 7.89 & 29.37 & \third{0.904} & \second{0.241} & 2.81 & 107 
        & 5.78 & 29.21 & 0.893 & 0.251 & 1.66 & 139 \\
        DashGaussian 
        & 4.54 & 29.18 & 0.903 & 0.253 & 2.21 & 109 
        & 3.82 & 30.02 & 0.908 & 0.259 & 1.07 & 156 \\
        FastGS 
        & \second{3.39} & \best{29.74} & \best{0.907} & \third{0.244} & 0.71 & \third{265}
        & \second{3.29} & \best{30.75} & \best{0.915} & \best{0.237} & 0.58 & \second{210} \\
        Ours (Taming) 
        & \third{3.72} & 29.06 & 0.897 & 0.246 & 1.23/1.33 & 120
        & \third{3.49} & 29.69 & 0.902 & \second{0.240} & 0.82/1.06 & 134 \\
        Ours (FastGS) 
        & \best{2.96} & \second{29.60} & 0.901 & 0.266 & 0.42/0.26 & 182
        & \best{3.12} & \second{30.73} & \second{0.913} & 0.248 & 0.32/0.33 & 164 \\
        
        \bottomrule
    \end{tabular}
    }
\end{table*}

\begin{table*}[t]
    \centering
    \caption{Results of the \emph{1K} setting. Scene-wise quantitative results over Tanks \& Temples dataset. Time is measured in minutes. $N_{\mathrm{GS}}$ denotes the number of Gaussians (in millions).}
    \label{tab:1K tanks&temples scene-wise results}
    \small
    \setlength{\tabcolsep}{4pt}
    \renewcommand{\arraystretch}{1.05}

    \resizebox{!}{0.9\height}{
    \begin{tabular}{lcccccc cccccc}
        \toprule
        \multirow{2}{*}{Method}
        & \multicolumn{6}{c}{\textbf{train}} 
        & \multicolumn{6}{c}{\textbf{truck}} \\
        \cmidrule(lr){2-7} \cmidrule(lr){8-13}
        & Time$\downarrow$ & PSNR$\uparrow$ & SSIM$\uparrow$ & LPIPS$\downarrow$
        & $N_{\mathrm{GS}}(M)\downarrow$ & FPS$\uparrow$
        & Time$\downarrow$ & PSNR$\uparrow$ & SSIM$\uparrow$ & LPIPS$\downarrow$
        & $N_{\mathrm{GS}}(M)\downarrow$ & FPS$\uparrow$ \\
        \midrule

        Original 3DGS 
        & 7.65  & 22.17  & \third{0.821}  & \second{0.196}  & 1.09  & 159 
        & 11.29  & 25.47  & \third{0.885}  & \third{0.142}  & 2.06 & 154  \\
        Opti3DGS
        & 8.83 & 21.69 & 0.806 & 0.221 & 0.78 & \third{218}
        & 11.62 & 25.47 & 0.879 & 0.161 & 1.68 & 130 \\
        Mini-Splatting 
        & 10.64  & 21.47  & 0.797  & 0.246  & 0.19 & \best{338} 
        & 10.45  & 25.08  & 0.873  & 0.160  & 0.21 & \second{161} \\
        Speedy-Splat 
        & 7.91  & 21.71  & 0.773  & 0.291  & 0.11 & 192 
        & 9.85  & 25.12  & 0.867  & 0.190  & 0.25 & 146  \\
        Taming-3DGS 
        & 5.83 & \best{22.80} & \best{0.832} & \best{0.189} & 1.10 & 147 
        & 6.83 & \second{25.93} & \second{0.887} & \second{0.141} & 1.89 & 136 \\
        DashGaussian 
        & 4.37 & \third{22.46} & 0.819 & 0.209 & 0.94 & 140
        & \third{4.07} & \third{25.83} & 0.881 & 0.162 & 1.36 & \third{156} \\
        FastGS 
        & \second{3.35} & \second{22.58} & \second{0.828} & \third{0.207} & 0.45 & \second{235}
        & \second{3.52} & \best{26.10} & \best{0.889} & \best{0.140} & 0.61 & \best{180} \\
        Ours (Taming) 
        & \third{3.52} & 22.28 & 0.801 & 0.239 & 0.47/0.47 & 110
        & 4.30 & 25.59 & 0.880 & 0.155 & 0.67/1.29 & 109 \\
        Ours (FastGS) 
        & \best{3.19} & 22.43 & 0.805 & 0.244 & 0.22/0.26 & 130
        & \best{3.07} & 25.63 & 0.880 & 0.166 & 0.26/0.46 & 118 \\
        
        \bottomrule
    \end{tabular}
    }
\end{table*}

\begin{table*}[t]
    \centering
    \caption{Results of the 2K setting. Scene-wise quantitative results over Mip-NeRF 360 dataset. Time is measured in minutes. $N_{\mathrm{GS}}$ denotes the number of Gaussians (in millions).}
    \label{tab:2K mip-nerf 360 scene-wise results}
    \small
    \setlength{\tabcolsep}{4pt}
    \renewcommand{\arraystretch}{1.05}
  
    \resizebox{\textwidth}{!}{ 
    \begin{tabular}{lcccccc cccccc cccccc}
        \toprule
        \multirow{2}{*}{Method}
        & \multicolumn{6}{c}{\textbf{bicycle}} & \multicolumn{6}{c}{\textbf{bonsai}} & \multicolumn{6}{c}{\textbf{counter}} \\
        \cmidrule(lr){2-7} \cmidrule(lr){8-13} \cmidrule(lr){14-19}
        & Time$\downarrow$ & PSNR$\uparrow$ & SSIM$\uparrow$ & LPIPS$\downarrow$ & $N_{\mathrm{GS}}(M)\!\downarrow$ & FPS$\uparrow$ & Time$\downarrow$ & PSNR$\uparrow$ & SSIM$\uparrow$ & LPIPS$\downarrow$ & $N_{\mathrm{GS}}(M)\!\downarrow$ & FPS$\uparrow$ & Time$\downarrow$ & PSNR$\uparrow$ & SSIM$\uparrow$ & LPIPS$\downarrow$ & $N_{\mathrm{GS}}(M)\!\downarrow$ & FPS$\uparrow$ \\
        \midrule
        Taming-3DGS
        & 13.18 & 24.10 & 0.673 & 0.394 & 2.87 & 47 
        & 7.02 & \second{32.47} & \second{0.940} & \third{0.244} & 1.04 & 91
        & 7.92 & \third{29.24} & \second{0.912} & \second{0.241} & 0.94 & 73
        \\
        DashGaussian 
        & \third{6.51} & 23.88 & 0.665 & 0.398 & 2.76 & 38
        & \third{4.70} & 31.72 & 0.933 & 0.255 & 0.78 & 78
        & 4.88 & 28.77 & 0.904 & 0.259 & 0.71 & 79
        \\
        FastGS   
        & \second{6.44} & \second{24.69} & \second{0.727} & \second{0.311} & 2.10 & \best{108}
        & 4.96 & \best{32.71} & \best{0.945} & \best{0.223} & 1.08 & \best{153}
        & \second{4.41} & \best{29.48} & \best{0.916} & \best{0.230} & 0.56 & \second{126}
        \\
        EfficientGS 
        & 42.72 & \best{24.73} & \best{0.735} & \best{0.293} & 2.18 & 53
        & 19.25 & 31.53 & 0.933 & 0.253 & 0.36 & \third{121}
        & 23.19 & 28.65 & 0.903 & 0.257 & 0.30 & 83
        \\
        Ours (Taming)  
        & 7.24 & 23.85 & 0.662 & 0.392 & 1.52/1.51 & \third{54}
        & \third{4.69} & 32.13 & 0.935 & 0.254 & 0.89/0.58 & 97
        & \third{4.81} & 29.14 & 0.905 & 0.257 & 0.62/0.43 & \third{94}
        \\
        Ours (FastGS)  
        & \best{5.52} & \third{24.46} & \third{0.702} & \third{0.337} & 2.31/0.88 & \second{104}
        & \best{4.17} & \third{32.28} & \third{0.940} & \second{0.236} & 1.01/0.42 & \second{136}
        & \best{3.82} & \second{29.39} & \third{0.910} & \third{0.247} & 0.50/0.23 & \best{132}
        \\
        \midrule
        \midrule
        \multirow{2}{*}{Method}
        & \multicolumn{6}{c}{\textbf{flowers}} & \multicolumn{6}{c}{\textbf{garden}} & \multicolumn{6}{c}{\textbf{kitchen}} \\
        \cmidrule(lr){2-7} \cmidrule(lr){8-13} \cmidrule(lr){14-19}
        & Time$\downarrow$ & PSNR$\uparrow$ & SSIM$\uparrow$ & LPIPS$\downarrow$ & $N_{\mathrm{GS}}(M)\!\downarrow$ & FPS$\uparrow$ & Time$\downarrow$ & PSNR$\uparrow$ & SSIM$\uparrow$ & LPIPS$\downarrow$ & $N_{\mathrm{GS}}(M)\!\downarrow$ & FPS$\uparrow$ & Time$\downarrow$ & PSNR$\uparrow$ & SSIM$\uparrow$ & LPIPS$\downarrow$ & $N_{\mathrm{GS}}(M)\!\downarrow$ & FPS$\uparrow$ \\
        \midrule
        Taming-3DGS
        & 11.30 & 20.45 & 0.540 & 0.453 & 2.00 & 53 
        & 15.06 & 26.48 & 0.794 & 0.232 & 3.84 & 44
        & 9.98 & \third{31.57} & \third{0.923} & \third{0.164} & 1.54 & 73
        \\
        DashGaussian 
        & \second{6.11} & 20.05 & 0.517 & 0.476 & 1.96 & 48 
        & \second{6.85} & 25.99 & 0.763 & 0.284 & 2.88 & 44
        & \best{5.02} & 31.22 & 0.915 & 0.181 & 1.02 & 83
        \\
        FastGS   
        & \third{6.27} & \second{20.97} & \second{0.564} & \second{0.423} & 1.62 & \second{97} 
        & 11.92 & \third{26.55} & \best{0.817} & \best{0.180} & 5.18 & \best{99}
        & 6.42 & \best{32.17} & \best{0.932} & \best{0.146} & 1.43 & \best{164}
        \\
        EfficientGS 
        & 40.76 & \best{21.20} & \best{0.580} & \best{0.392} & 1.77 & 55
        & 47.52 & \second{26.58} & \third{0.808} & \second{0.210} & 2.25 & \third{58}
        & 25.09 & 31.07 & 0.918 & 0.172 & 0.50 & \third{100}
        \\
        Ours (Taming)  
        & 6.93 & 20.38 & 0.526 & 0.463 & 1.13/1.18 & \third{61} 
        & \third{8.56} & 26.33 & 0.783 & 0.249 & 1.62/2.35 & 45
        & \third{5.37} & 31.13 & 0.916 & 0.184 & 0.88/0.51 & 84
        \\
        Ours (FastGS)  
        & \best{5.18} & \third{20.82} & \third{0.546} & \third{0.441} & 1.32/0.68 & \best{101}
        & \best{6.82} & \best{26.72} & \second{0.809} & \third{0.212} & 2.23/2.25 & \second{74}
        & \second{5.03} & \second{32.13} & \second{0.930} & \second{0.155} & 1.14/0.71 & \second{139}
        \\
        \midrule
        \midrule
        \multirow{2}{*}{Method}
        & \multicolumn{6}{c}{\textbf{room}} & \multicolumn{6}{c}{\textbf{stump}} & \multicolumn{6}{c}{\textbf{treehill}} \\
        \cmidrule(lr){2-7} \cmidrule(lr){8-13} \cmidrule(lr){14-19}
        & Time$\downarrow$ & PSNR$\uparrow$ & SSIM$\uparrow$ & LPIPS$\downarrow$ & $N_{\mathrm{GS}}(M)\!\downarrow$ & FPS$\uparrow$ & Time$\downarrow$ & PSNR$\uparrow$ & SSIM$\uparrow$ & LPIPS$\downarrow$ & $N_{\mathrm{GS}}(M)\!\downarrow$ & FPS$\uparrow$ & Time$\downarrow$ & PSNR$\uparrow$ & SSIM$\uparrow$ & LPIPS$\downarrow$ & $N_{\mathrm{GS}}(M)\!\downarrow$ & FPS$\uparrow$ \\
        \midrule
        Taming-3DGS
        & 7.68 & \second{31.63} & 0.916 & 0.266 & 1.08 & 83
        & 10.12 & 25.76 & 0.751 & 0.379 & 2.39 & 49
        & 11.11 & \best{22.55} & \second{0.636} & \second{0.441} & 1.67 & 48
        \\
        DashGaussian 
        & 4.74 & 31.39 & 0.911 & 0.278 & 0.81 & 75 
        & \third{5.84} & 25.42 & 0.734 & 0.402 & 2.49 & 48
        & \third{6.28} & 22.42 & 0.624 & 0.456 & 1.92 & 45
        \\
        FastGS   
        & \third{4.15} & \best{31.86} & \best{0.923} & \best{0.244} & 0.71 & \best{139} 
        & \second{5.57} & \second{26.90} & \second{0.785} & \second{0.332} & 1.25 & \best{108}
        & \second{5.46} & \third{22.42} & \third{0.633} & \third{0.452} & 1.11 & \best{92}
        \\
        EfficientGS 
        & 22.00 & 31.03 & \third{0.918} & \second{0.256} & 0.40 & \third{108}
        & 36.95 & \best{26.90} & \best{0.790} & \best{0.318} & 1.56 & 48
        & 42.69 & 21.92 & \best{0.643} & \best{0.396} & 2.03 & 58
        \\
        Ours (Taming)  
        & \second{4.14} & 31.20 & 0.911 & 0.276 & 0.60/0.46 & 102 
        & 6.62 & 25.31 & 0.730 & 0.394 & 1.60/1.31 & \third{54}
        & 7.13 & 22.40 & 0.618 & 0.453 & 1.08/1.00 & \third{59}
        \\
        Ours (FastGS)  
        & \best{3.55} & \third{31.56} & \second{0.920} & \third{0.256} & 0.54/0.28 & \second{120}
        & \best{4.78} & \third{26.61} & \third{0.770} & \third{0.348} & 1.30/0.52 & \second{81}
        & \best{4.49} & \second{22.43} & 0.623 & 0.465 & 0.66/0.34 & \second{79}
        \\
        \bottomrule
    \end{tabular}
    } 
\end{table*}

\begin{table*}[t]
    \centering
    \caption{Results of the \emph{4K} setting. Scene-wise quantitative results over Mip-NeRF 360 dataset. Time is measured in minutes. $N_{\mathrm{GS}}$ denotes the number of Gaussians (in millions).}
    \label{tab:4K mip-nerf 360 scene-wise results}
    \small
    \setlength{\tabcolsep}{4pt}
    \renewcommand{\arraystretch}{1.05}
  
    \resizebox{\textwidth}{!}{ 
    \begin{tabular}{lcccccc cccccc cccccc}
        \toprule
        \multirow{2}{*}{Method}
        & \multicolumn{6}{c}{\textbf{bicycle}} & \multicolumn{6}{c}{\textbf{bonsai}} & \multicolumn{6}{c}{\textbf{counter}} \\
        \cmidrule(lr){2-7} \cmidrule(lr){8-13} \cmidrule(lr){14-19}
        & Time$\downarrow$ & PSNR$\uparrow$ & SSIM$\uparrow$ & LPIPS$\downarrow$ & $N_{\mathrm{GS}}(M)\!\downarrow$ & FPS$\uparrow$ & Time$\downarrow$ & PSNR$\uparrow$ & SSIM$\uparrow$ & LPIPS$\downarrow$ & $N_{\mathrm{GS}}(M)\!\downarrow$ & FPS$\uparrow$ & Time$\downarrow$ & PSNR$\uparrow$ & SSIM$\uparrow$ & LPIPS$\downarrow$ & $N_{\mathrm{GS}}(M)\!\downarrow$ & FPS$\uparrow$ \\
        \midrule
        Taming-3DGS
        & 23.28 & 23.66 & 0.686 & 0.441 & 2.19 & 29 
        & 9.80 & \second{32.07} & \second{0.935} & \third{0.293} & 0.96 & 72 
        & 11.21 & 28.95 & \third{0.910} & \third{0.286} & 0.88 & 65 
        \\
        DashGaussian 
        & \third{9.92} & 23.52 & 0.683 & 0.445 & 2.16 & 25 
        & \second{6.02} & 31.49 & 0.930 & 0.300 & 0.78 & 64 
        & \third{6.37} & 28.50 & 0.904 & 0.298 & 0.69 & 56 
        \\
        FastGS   
        & \second{9.55} & \second{24.53} & \second{0.734} & \third{0.366} & 1.92 & \best{109} 
        & \third{6.20} & \best{32.74} & \best{0.942} & \best{0.271} & 1.24 & \best{122} 
        & \second{5.56} & \best{29.43} & \best{0.916} & \best{0.270} & 0.60 & \second{129}
        \\
        EfficientGS 
        & 91.55 & \best{24.66} & \best{0.743} & \best{0.347} & 2.04 & \third{36} 
        & 33.48 & 31.01 & 0.930 & 0.298 & 0.37 & \second{96}
        & 39.86 & 28.52 & 0.904 & 0.294 & 0.31 & 65 
        \\
        Ours (Taming)  
        & 12.16 & 23.65 & 0.683 & 0.437 & 1.25/1.06 & 35
        & 6.50 & 31.91 & 0.930 & 0.301 & 0.88/0.53 & 65
        & 6.68 & \third{29.03} & 0.905 & 0.297 & 0.62/0.40 & \third{67}
        \\
        Ours (FastGS)  
        & \best{8.38} & \third{24.43} & \third{0.729} & \second{0.358} & 2.78/0.68 & \second{83}
        & \best{4.94} & \third{31.98} & \third{0.935} & \second{0.284} & 1.22/0.41 & \third{78}
        & \best{4.55} & \second{29.39} & \second{0.912} & \second{0.286} & 0.58/0.20 & \best{144}
        \\
        \midrule
        \midrule
        \multirow{2}{*}{Method}
        & \multicolumn{6}{c}{\textbf{flowers}} & \multicolumn{6}{c}{\textbf{garden}} & \multicolumn{6}{c}{\textbf{kitchen}} \\
        \cmidrule(lr){2-7} \cmidrule(lr){8-13} \cmidrule(lr){14-19}
        & Time$\downarrow$ & PSNR$\uparrow$ & SSIM$\uparrow$ & LPIPS$\downarrow$ & $N_{\mathrm{GS}}(M)\!\downarrow$ & FPS$\uparrow$ & Time$\downarrow$ & PSNR$\uparrow$ & SSIM$\uparrow$ & LPIPS$\downarrow$ & $N_{\mathrm{GS}}(M)\!\downarrow$ & FPS$\uparrow$ & Time$\downarrow$ & PSNR$\uparrow$ & SSIM$\uparrow$ & LPIPS$\downarrow$ & $N_{\mathrm{GS}}(M)\!\downarrow$ & FPS$\uparrow$ \\
        \midrule
        Taming-3DGS
        & 21.16 & 20.07 & 0.567 & 0.486 & 1.52 & 33 
        & 25.96 & 26.04 & 0.774 & 0.310 & 3.20 & 29 
        & 13.86 & \third{31.22} & \third{0.918} & \third{0.204} & 1.45 & 61
        \\
        DashGaussian 
        & \second{9.58} & 19.62 & 0.552 & 0.506 & 1.54 & 28 
        & \second{10.83} & 25.63 & 0.754 & 0.339 & 2.69 & 26 
        & \second{6.61} & 30.83 & 0.911 & 0.219 & 0.98 & 59 
        \\
        FastGS   
        & \third{9.66} & \second{20.76} & \second{0.587} & \second{0.461} & 1.57 & \best{107} 
        & 17.47 & \best{26.60} & \best{0.810} & \best{0.227} & 5.93 & \best{80} 
        & 7.94 & \second{31.94} & \best{0.928} & \best{0.182} & 1.53 & \best{113} 
        \\
        EfficientGS 
        & 89.80 & \best{21.19} & \best{0.606} & \best{0.426} & 1.76 & \third{48} 
        & 104.86 & \third{26.41} & \second{0.802} & \second{0.260} & 2.39 & \third{41} 
        & 42.52 & 31.02 & 0.915 & 0.209 & 0.53 & \third{87} 
        \\
        Ours (Taming)  
        & 12.22 & 20.01 & 0.561 & 0.493 & 0.97/0.86 & 37
        & \third{14.16} & 26.02 & 0.770 & 0.307 & 1.67/1.76 & 33
        & \third{7.42} & 31.16 & 0.913 & 0.224 & 0.91/0.46 & 61
        \\
        Ours (FastGS)  
        & \best{7.62} & \third{20.53} & \third{0.577} & \third{0.466} & 1.53/0.52 & \second{94}
        & \second{9.69} & \second{26.47} & \third{0.798} & \third{0.266} & 3.15/1.51 & \second{71}
        & \best{5.74} & \best{31.99} & \second{0.926} & \second{0.196} & 1.35/0.56 & \second{110}
        \\
        \midrule
        \midrule
        \multirow{2}{*}{Method}
        & \multicolumn{6}{c}{\textbf{room}} & \multicolumn{6}{c}{\textbf{stump}} & \multicolumn{6}{c}{\textbf{treehill}} \\
        \cmidrule(lr){2-7} \cmidrule(lr){8-13} \cmidrule(lr){14-19}
        & Time$\downarrow$ & PSNR$\uparrow$ & SSIM$\uparrow$ & LPIPS$\downarrow$ & $N_{\mathrm{GS}}(M)\!\downarrow$ & FPS$\uparrow$ & Time$\downarrow$ & PSNR$\uparrow$ & SSIM$\uparrow$ & LPIPS$\downarrow$ & $N_{\mathrm{GS}}(M)\!\downarrow$ & FPS$\uparrow$ & Time$\downarrow$ & PSNR$\uparrow$ & SSIM$\uparrow$ & LPIPS$\downarrow$ & $N_{\mathrm{GS}}(M)\!\downarrow$ & FPS$\uparrow$ \\
        \midrule
        Taming-3DGS
        & 10.84 & \best{31.49} & 0.912 & 0.306 & 1.04 & 60 
        & 18.40 & 25.51 & 0.779 & 0.438 & 1.72 & 33 
        & 21.79 & \second{22.31} & \second{0.673} & \third{0.481} & 1.20 & 27 
        \\
        DashGaussian 
        & 6.42 & 31.20 & 0.908 & 0.312 & 0.80 & 51 
        & \third{9.81} & 25.04 & 0.766 & 0.457 & 1.79 & 28 
        & \third{9.80} & 22.24 & 0.665 & 0.490 & 1.42 & 25 
        \\
        FastGS   
        & \second{5.28} & \third{31.35} & \best{0.917} & \best{0.282} & 0.77 & \second{115} 
        & \second{8.44} & \second{26.71} & \second{0.806} & \third{0.395} & 1.09 & \best{94} 
        & \second{8.54} & \third{22.26} & \third{0.671} & 0.485 & 0.88 & \second{78} 
        \\
        EfficientGS 
        & 37.69 & 30.98 & \third{0.915} & \second{0.293} & 0.42 & \third{99} 
        & 79.42 & \best{26.88} & \best{0.812} & \best{0.376} & 1.47 & \third{39} 
        & 93.48 & 22.01 & \best{0.684} & \best{0.440} & 1.87 & \third{33} 
        \\
        Ours (Taming)  
        & \third{5.53} & 31.05 & 0.909 & 0.312 & 0.61/0.40 & 71
        & 11.85 & 25.08 & 0.767 & 0.444 & 1.26/0.99 & 35
        & 13.08 & \best{22.41} & 0.669 & \second{0.480} & 0.84/0.69 & 32
        \\
        Ours (FastGS)  
        & \best{4.17} & \second{31.43} & \second{0.916} & \third{0.297} & 0.65/0.20 & \best{143}
        & \best{7.20} & \third{26.55} & \third{0.801} & \second{0.392} & 1.33/0.46 & \second{58}
        & \best{6.81} & 22.22 & 0.670 & 0.487 & 0.63/0.25 & \best{78}
        \\
        \bottomrule
    \end{tabular}
    } 
\end{table*}

\clearpage

\begin{table}[h!]
    \centering
    \caption{Ablation studies over the number of Laplacian levels on the Mip-NeRF
360 dataset. (\emph{1K} setting)}
    \label{tab:the number of Laplacian levels, 1k}
    \small
    \setlength{\tabcolsep}{4pt}
    \renewcommand{\arraystretch}{1.05}
    \begin{tabular}{llcccccc}
        \toprule
        Method & Iteration Split & Time$\downarrow$ & PSNR$\uparrow$ & SSIM$\uparrow$ & LPIPS$\downarrow$ & $N_{\mathrm{GS}}(M)\!\downarrow$ & FPS$\uparrow$ \\
        \midrule
        Two-level & (10000, 20000)
        & 3.96 & \textbf{27.67} & \textbf{0.809} & \textbf{0.233} & 0.71/0.87 & \textbf{122} \\
        Three-level & (10000, 5000, 15000)
        & 3.59 & 27.37 & 0.796 & 0.255 & 0.31/0.39/0.82 & 107 \\
        Four-level & (10000,5000,5000,10000)
        & \textbf{3.31} & 26.90 & 0.781 & 0.274 & 0.13/0.18/0.33/0.71 & 89 \\
        \bottomrule
    \end{tabular}
\end{table}

\begin{table*}[h!]
    \centering

    \caption{Effects of Laplacian levels and training iterations under the \emph{4K} setting on the Mip-NeRF 360 dataset. We report results with 30k iterations, and additionally with 45k iterations to improve convergence. Both settings use the same split ratios across levels.}

    \label{tab:more iter./levels results on 4K dataset}
    \small
    \setlength{\tabcolsep}{4pt}
    \renewcommand{\arraystretch}{1.05}
    \begin{tabular}{lllcccccc}
        \toprule
        Levels & Iteration & Iteration Split & Time$\downarrow$ & PSNR$\uparrow$ & SSIM$\uparrow$ & LPIPS$\downarrow$ & $N_{\mathrm{GS}}(M)\!\downarrow$ & FPS$\uparrow$ \\
        \midrule
        Two-level & 30k & (10000, 20000)
        & 6.57 & \textbf{27.22} & \textbf{0.807} & \textbf{0.337} & 1.47/0.53 & 95 \\
        Three-level & 30k & (10000, 5000, 15000)
        & 5.99 & 27.17 & 0.801 & 0.358 & 0.94/0.69/0.40 & \textbf{97} \\
        Four-level & 30k & (10000, 5000, 5000, 10000)
        & \textbf{5.09} & 26.84 & 0.792 & 0.376 & 0.45/0.46/0.48/0.35 & 74 \\
        \midrule
        Two-level & 45k & (15000, 30000)
        & 10.61 & \textbf{27.39} & \textbf{0.810} & \textbf{0.331} & 1.74/0.54 & \textbf{98} \\
        Three-level & 45k & (15000, 7500, 22500)
        & 8.65 & 27.27 & 0.803 & 0.353 & 1.07/0.63/0.38 & 88 \\
        Four-level & 45k & (15000, 7500, 7500, 15000)
        & \textbf{7.54} & 26.97 & 0.795 & 0.370 & 0.49/0.46/0.47/0.35 & 80 \\
        \bottomrule
    \end{tabular}
\end{table*}

\end{document}